\documentclass[sigconf]{acmart}
\usepackage{multirow}
\usepackage{subfigure}
\usepackage{bm}
\usepackage{multirow}
\usepackage{subfigure}
\usepackage[normalem]{ulem}
\useunder{\uline}{\ul}{}
\usepackage{subfigure}
\usepackage{algorithm}
\usepackage[noend]{algpseudocode}
\usepackage{cite}
\usepackage{enumitem}
\usepackage{bbm}

\AtBeginDocument{%
  \providecommand\BibTeX{{%
    \normalfont B\kern-0.5em{\scshape i\kern-0.25em b}\kern-0.8em\TeX}}}

\setcopyright{acmlicensed}
\copyrightyear{2018}
\acmYear{2018}
\acmDOI{XXXXXXX.XXXXXXX}

\acmConference[Conference acronym 'XX]{Make sure to enter the correct
  conference title from your rights confirmation emai}{June 03--05,
  2018}{Woodstock, NY}
\acmISBN{978-1-4503-XXXX-X/18/06}

\begin{document}

\title{Large Language Model for Participatory Urban Planning}

\author{Zhilun Zhou}
\affiliation{%
  \institution{Department of Electronic Engineering,  Tsinghua University}
  \city{Beijing}
  \country{China}
}
\email{zzl22@mails.tsinghua.edu.cn}

\author{Yuming Lin}
\authornote{Corresponding Author.}
\orcid{0000-0003-0442-7071}
\affiliation{%
  \institution{Department of Electronic Engineering, Tsinghua University}
  \city{Beijing}
  \country{China}
}
\email{linyuming9@mail.tsinghua.edu.cn}

\author{Depeng Jin}
\orcid{0000-0003-0419-5514}
\affiliation{%
  \institution{Department of Electronic Engineering, Tsinghua University}
  \city{Beijing}
  \country{China}
}
\email{jindp@tsinghua.edu.cn}

\author{Yong Li}
\orcid{0000-0001-5617-1659}
\affiliation{%
  \institution{Department of Electronic Engineering, Tsinghua University}
  \city{Beijing}
  \country{China}
}
\email{liyong07@tsinghua.edu.cn}

\renewcommand{\shortauthors}{Trovato and Tobin, et al.}

\begin{abstract}
Participatory urban planning is the mainstream of modern urban planning that involves the active engagement of residents. However, the traditional participatory paradigm requires experienced planning experts and is often time-consuming and costly. Fortunately, the emerging Large Language Models (LLMs) have shown considerable ability to simulate human-like agents, which can be used to emulate the participatory process easily. In this work, we introduce an LLM-based multi-agent collaboration framework for participatory urban planning, which can generate land-use plans for urban regions considering the diverse needs of residents. Specifically, we construct LLM agents to simulate a planner and thousands of residents with diverse profiles and backgrounds. We first ask the planner to carry out an initial land-use plan. To deal with the different facilities needs of residents, we initiate a discussion among the residents in each community about the plan, where residents provide feedback based on their profiles. Furthermore, to improve the efficiency of discussion, we adopt a fishbowl discussion mechanism, where part of the residents discuss and the rest of them act as listeners in each round. Finally, we let the planner modify the plan based on residents' feedback. We deploy our method on two real-world regions in Beijing. Experiments show that our method achieves state-of-the-art performance in residents satisfaction and inclusion metrics, and also outperforms human experts in terms of service accessibility and ecology metrics.

\end{abstract}






\maketitle

\section{Introduction}


Participatory urban planning is a paradigm of urban planning, characterized by collaborative decision-making that involves the active engagement of different stakeholders, including officials, professionals, developers, local residents, and public~\citep{arnstein1969ladder, forester1982planning, forester1999deliberative}. This inclusive approach aims to incorporate diverse perspectives in planning and decision-making processes through carefully designed procedures, discussions, panels, or workshops, making it one of the most commonly used methods in current community planning~\citep{li2020collaborative}.

Nevertheless, the traditional participatory urban planning method requires extensive experience, consideration of multiple interests, and often a case-by-case examination of unique community problems. It demands significant time and manpower and heavily relies on experienced urban planners~\citep{eriksson2022opening}. As cities continue to grow and the need for urban renewal steadily increases, the limitations of traditional urban planning methods become more apparent, prompting the exploration of innovative methodologies that can enhance their efficiency and effectiveness~\citep{tian2023participatory}.

On the other hand, various generative urban planning techniques have been proposed, such as evolutionary algorithms~\citep{koenig2020inteegrating}, Generative Adversarial Networks (GAN)~\citep{quan2022urban}, Variational Autoencoder (VAE)~\citep{wang2021deep}, or Reinforcement Learning (RL)~\citep{zheng2023spatial, qian2023ai}. These methods largely improve the efficiency of urban planning, while they fail to incorporate individual residents and consider their diverse demands, and thus are not "participatory" actually.


Large Language Models (LLMs) have emerged as a promising solution. They are now capable of creating human-like agents, which can mirror the interests and demands of people with simple role-play prompts~\citep{shanahan2023role,park2023generative}. This unique ability makes them effective representatives of diverse participants in urban planning scenarios. In this study, we introduce LLMs to participatory urban planning by crafting LLM agents to simulate the planner and residents, and generate land-use plans for urban regions through multi-agent collaboration.

However, to achieve effective and efficient planning performance, we still face two challenges: (1) How to balance the diverse needs of residents? Residents have various profiles and family backgrounds, resulting in different needs for land use in their neighborhoods. For example, families with patients may prefer medical services and parks nearby, while families with children may need school. Therefore, residents may have different opinions on lane-use choices, and it is difficult to comprehensively consider the diverse needs of residents. (2) How to achieve an efficient participatory planning process given the large number of residents? A typical urban region may consider thousands of residents, making it time-consuming to communicate with all residents. Moreover, there exists a limit of context length in most LLMs, which cannot support overlong dialogues. It is important but challenging to reduce the time cost and token cost.

To tackle these challenges, we propose a participatory planning framework through multi-agent collaboration. Specifically, the planner agent first proposes an initial land-use plan for the region based on expert knowledge. To deal with the diverse needs of residents, a discussion among neighboring resident agents is initiated in each community, where the residents communicate with each other about their opinions on the land-use plan. Finally, the planner revised the plan according to the feedback of residents. To address the second challenge, we adopt a fishbowl discussion mechanism where participants are divided into inner and outer circles. In each round, only participants in the inner circle discuss, and the rest of them listen to the discussion. After each round, members in the two circles are randomly exchanged. Moreover, the discussion history is summarized to avoid overlong contexts. 
Through such design, the planner and residents agents are able to collaboratively generate a land-use plan considering different residents' needs efficiently and effectively.

Overall, our contributions are summarized as follows:
\begin{itemize}
    \item To our best knowledge, we are the first to introduce LLMs to participatory urban planning by creating LLM agents to simulate the planning process. Our methods can not only generate reasonable and effective land-use plans but also allow the governor and planners to simulate the entire planning process before actual planning activities begin.
    \item We propose a planning framework incorporating residents' discussion and planner revision to better cater to residents' needs. Moreover, we adopt a fishbowl discussion mechanism to ensure more efficient discussion.
    \item We deploy our methods on two urban regions in Beijing, and the results show that our methods outperform human experts in satisfying residents' needs as well as service accessibility and ecology metrics. 
\end{itemize}

\section{Related Work}
\label{sec:related work}

\subsection{Participatory Planning}
Participatory Planning has evolved as a pivotal approach in urban planning, emphasizing community engagement in the decision-making process~\citep{forester1999deliberative,arnstein1969ladder,forester1982planning}. Previous studies have proposed various forms of activities to engage residents in community planning, like community mapping, World Cafe, Fishbowl Discussion, visioning workshop, etc.~\citep{liu2022participatory}. These
However, traditional methods have encountered many challenges in practice, including high cost, lack of skilled facilitators, low efficiency, and low interest to participate~\citep{abas2023systematic}. 
The recent advances in information technologies can help the practice of participatory democracy. For example, lots of digital participatory planning practices show that Augmented Reality (AR) and Virtual Reality (VR) can offer immersive and inclusive environments, facilitating diverse stakeholder participation~\citep{jiang2018demonstrator, sasmannshausen2021citizen, chassin2022experiencing, nasrazadani2022rapid,  ahmadi2023augmented}. Moreover, AI techniques can be helpful in participant selection, identifying domain experts, and weighing opinions, thereby easing the burden on facilitators and promoting efficiency~\citep{du2023artificial}.  
However, despite the promising potential of technological innovations to enhance citizen engagement, there remains a significant gap in truly transforming group processes and outcomes~\citep{du2023artificial}. The limited familiarity and a dearth of consistency and transparency restrain the more substantial integration of AI methods~\citep{lock2021towards}. The ongoing progress in natural language processing holds the promise of addressing these challenges and unlocking the full potential of technological interventions in participatory processes.

\subsection{Multi-agent Collaboration with Large Language Models}
There has been an adequate study showing that LLM can serve as an emulated role-play tool~\citep{shanahan2023role}. Many studies accompanying the recent advancement of LLM now focus on multi-agent collaboration with LLMs, where different roles will be assigned to LLM agents. Then, these agents will cooperate to solve a complex task together. 

Specifically, LLM agents are usually assigned roles based on distinct expert knowledge in the role-play part. For example, solving operations research problems needs roles like terminology interpreter, modeling expert, programmer, and evaluator~\citep{anonymous2024chainofexperts}, while developing software may need product manager, architect, project manager, engineer, and QA engineer~\citep{hong2023metagpt}. Moreover, some studies craft LLM agents with adaptive roles when solving the problem~\citep{chen2023agentverse}. To enable LLM agents to act in specific roles, researchers usually carefully design prompts to guide agents' behavior and provide expert knowledge needed by prompting or external knowledge bases.

After the role-play, collaboration mechanisms are designed to enable the cooperation of different agents. Such a mechanism can be the form of a pipeline, where different agents sequentially finish part of the task~\citep{hong2023metagpt,anonymous2024chainofexperts,yu2023thought}, or group discussion, where several agents communicate with each other to reach an agreement~\citep{wu2023autogen,zhang2023exploring}, or the hybrid of them~\citep{chen2023agentverse}. 

Multi-agent collaboration with LLMs has shown considerable success in many domains, including operations research~\citep{anonymous2024chainofexperts}, text quality evaluation~\citep{chan2023chateval}, solving math problems~\citep{li2023camel}, and software development~\citep{hong2023metagpt}. However, none of the existing studies have applied LLMs to simulate citizens in urban planning scenarios. Moreover, most existing works only incorporate a few LLM agents, which cannot be directly applied to participatory planning where a community may contain a dozen thousand residents.

\section{Problem Statement}
\label{sec:problem_statement}

\begin{figure*}[htbp!]
    \centering
    \includegraphics[width=.9\linewidth]{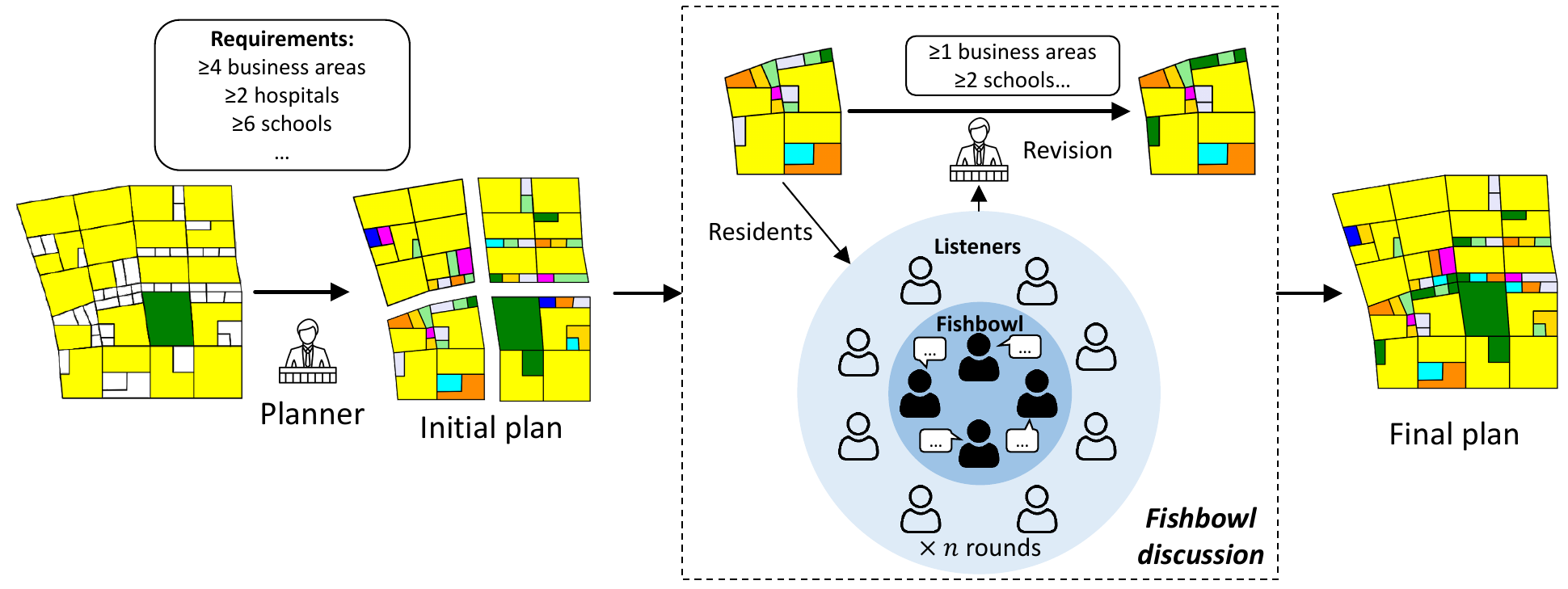}
    \caption{The framework of our proposed participatory planning method.}
    \label{fig:framework}
\end{figure*}

In this work, we investigate the task of region renovation~\citep{zheng2023spatial}, which is defined as follows.
\begin{definition}[\textbf{Region Renovation}]
Given an urban region, we partitioned it into small areas based on road networks and existing land-use conditions. We retain the residential areas and some areas with fixed land use and leave other areas as vacant spaces, denoted as $\mathcal{A}=\{A_1,A_2,\ldots,A_{n_a}\}$. Given the available land-use types $\mathcal{U}=\{u_1,u_2,\ldots,u_{n_u}\}$, the region renovation task aims to choose a specific land-use type for each vacant area, i.e., design a plan $P:\mathcal{A}\rightarrow \mathcal{U}$.
Moreover, the number of each land-use type should satisfy quantity requirements $\left\lvert \{A_i|P(A_i)=u_k\} \right\rvert\geq s_k, k=1,2,\ldots,n_u$.
The goal of region renovation is to make it convenient and comfortable for residents to live, work, and access various services they need.
\end{definition}


\section{Methods}
\label{sec:methods}

Based on the definition of region renovation, we propose a LLM-based framework to solve the problem. To enable LLMs to act as participants in participatory planning, we assign LLM agents with different roles, including a planner and multiple residents, through carefully designed prompts.
Moreover, we design a planning workflow where the planner and residents collaborate to generate the renovation plan for the region, as shown in Figure~\ref{fig:framework}. Specifically, the planner first proposes an initial lane-use plan and then revises the plan based on the opinions of residents through discussion.


\subsection{Agent Roles Design}
\label{sec:role_play}
\begin{figure}[htbp!]
    \centering
    \includegraphics[width=.95\linewidth]{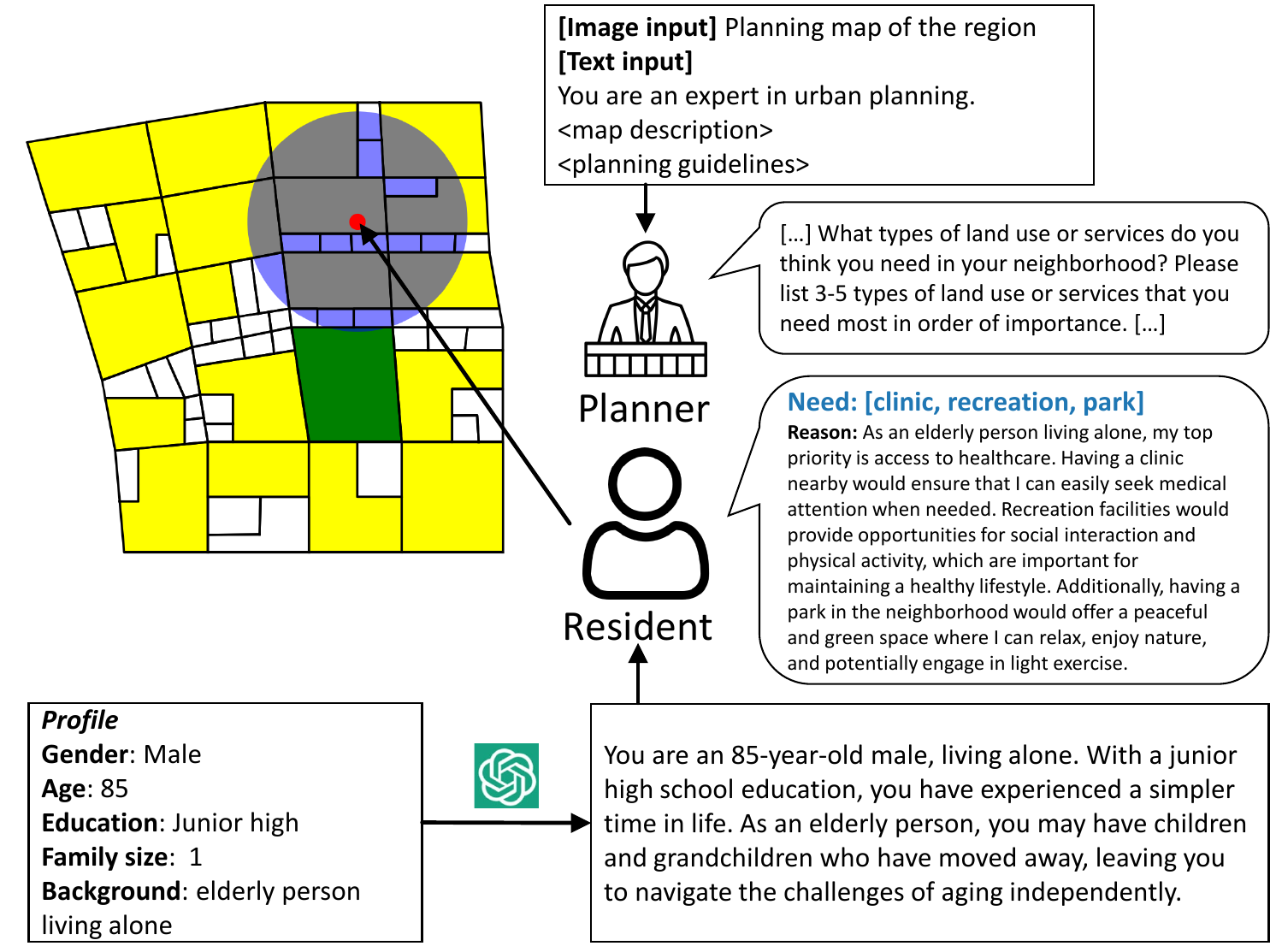}
    \caption{The design of planner and residents.}
    \label{fig:role_design}
\end{figure}

Leveraging the role-play ability of LLM, we construct diverse agents to simulate distinct real participants in urban planning, including a planner and multiple residents, as shown in Figure~\ref{fig:role_design}.

\subsubsection{Planner Design}
The planner is responsible for the overall planning considerations. To make the planner understand the spatial distribution of areas, we leverage the multi-modal capabilities of GPT-4v and input the map of the region, where the areas are labeled with IDs and filled with colors to represent different land uses. We also provide a textual description of the location of each area, including its position and nearby areas~\citep{li-etal-2023-theory}. Moreover, to enable the planner to generate a more realistic plan, we incorporate some planning guidelines designed by human experts in the prompts. We present all the prompts in Appendix~\ref{app:prompts}.

\subsubsection{Resident Design}
The most important part of participatory urban planning is the active engagement of residents in the planning process. Therefore, we design LLM agents to simulate the residents living in this region. Specifically, we generate the profile of each resident from the real profile distribution, including gender, age, education level, and family size. Moreover, to ensure fairness and inclusion, we add special backgrounds to some of the residents like parenting families, families with patients, and elderly persons living alone.
To further enhance the realness and diversity of residents, we ask the GPT to generate a short description based on the profile of each resident. 
Each resident is then assigned a home address randomly sampled from residential areas. Based on the idea of the 15-minute city, we let the residents only observe the areas near their homes within a 15-minute walk or cycling (set as 500m). Specifically, each resident is informed of the direction and distance of each area relative to his home address.

Residents with different profiles and family backgrounds may have different needs for land use and facilities in their neighborhood, which are essential to consider in participatory planning. Therefore, to elicit residents' needs, we directly ask them to list several types of land use they need most. As shown in Figure~\ref{fig:role_design}, residents can provide reasonable responses corresponding to their profiles. Moreover, the needs of residents are further used to measure residents' satisfaction with the land-use plan, which will be introduced in Section~\ref{sec:metrics}.

In summary, equipped with expert knowledge, the planner can observe the whole region from a global view, controlling the overall distribution of facilities. On the other hand, each resident focuses on his own needs and areas in the neighborhood, providing suggestions from a local view to improve the plan.
Such role design incorporates various stakeholders in real-world scenarios, enabling a more comprehensive view of participatory planning.

\subsection{Planning Workflow}

With the agents acting as the planner and residents, we propose a planning procedure through multi-agent collaboration, as shown in Figure~\ref{fig:framework}.
Particularly, the planner first proposes an initial plan based on the requirements and expert knowledge, including the land use for each area in the region. Then, the key idea of participatory planning is to ask the opinions of residents on this plan and revise it accordingly.

However, the challenge here is that different residents may have different needs for land use, leading to potential conflicts of interest among neighboring residents. To better balance their needs, we propose to let residents communicate with each other through a discussion. 
On the other hand, two residents living far away from each other will not have conflicts of interest, since residents only care about the areas near their homes. Therefore, the discussion does not necessarily need to include all residents, and it only needs to be conducted among neighboring residents instead.

Therefore, we partition the entire region into four distinct communities based on real-world communities and road networks. Subsequently, for each community, the residents living there or nearby (within a 500m buffer) are invited to discuss for several rounds, where they share opinions about the land-use plan. Based on residents' opinions in the discussion, the planner revises the plan of this community accordingly. 


\subsection{Fishbowl Discussion}
In the aforementioned discussion, another important problem is that the number of residents is very large. In China, a typical community often consists of tens of thousands of people. Even if only 1\% of them participate in the discussion, it would still lead to an overlong context and substantial time costs.
To address this challenge, we propose implementing the fishbowl discussion mechanism, which is commonly used in participatory planning~\citep{liu2022participatory}, to enhance the efficiency of discussion. Specifically, in each round, residents are grouped into inner and outer circles. The inner circle, often referred to as the "fishbowl," actively engages in discussions about the plan, while those in the outer circle listen attentively. Following each round of discussion, there's a role-switching process wherein some listeners join the fishbowl as speakers, and some previous speakers take on the role of listeners.
Moreover, to avoid an overlong context, the discussion is summarized after each round, and residents receive the summary of the previous discussion.

The whole discussion and renovation process for each community is presented in Algorithm~\ref{alg:discussion}.
Through such design, residents can exchange opinions with each other for multiple rounds to better balance their needs. Furthermore, the discussion can scale up to accommodate a large number of residents without significant time costs.

\algrenewcommand\algorithmicprocedure{\textbf{Step}}
\algnewcommand{\LineComment}[1]{\State \(\triangleright\) #1}
\begin{algorithm}
\caption{Community renovation by fishbowl discussion}\label{alg:discussion}
\begin{algorithmic}[1]
\State \textbf{Input}: Initial plan of the community $P_0$, the set of residents living in the community or nearby $\mathcal{R}_c$, planner
\State \textbf{Output}: Revised plan $P_1$
\State Initialize discussion history $H\leftarrow \emptyset$
\For{$round\in\{1,2,\ldots,N\}$}
    \State Sample $M$ residents $\{R_1,\ldots,R_M\}\in\mathcal{R}_c$ to discuss
    \For{$Resident\in\{R_1,\ldots,R_M\}$}
        \State $R_j$ observes the land-use in his neighborhood
        \State $R_j$ expresses his opinions $O_j$ about the plan
    \EndFor
    \State $S=Summarize(\{O_1,\ldots,O_M\})$
    \State Other residents receive $S$
    \State $H\leftarrow H\cup S$
\EndFor
\State Planner revise the plan based on $H$ to obtain $P_1$
\end{algorithmic}

\end{algorithm}

\section{Experiments}
\label{sec:experiments}

\subsection{Experiment Settings}
\subsubsection{Datasets}

\begin{figure}[htbp!]
    \centering
    \includegraphics[width=.9\linewidth]{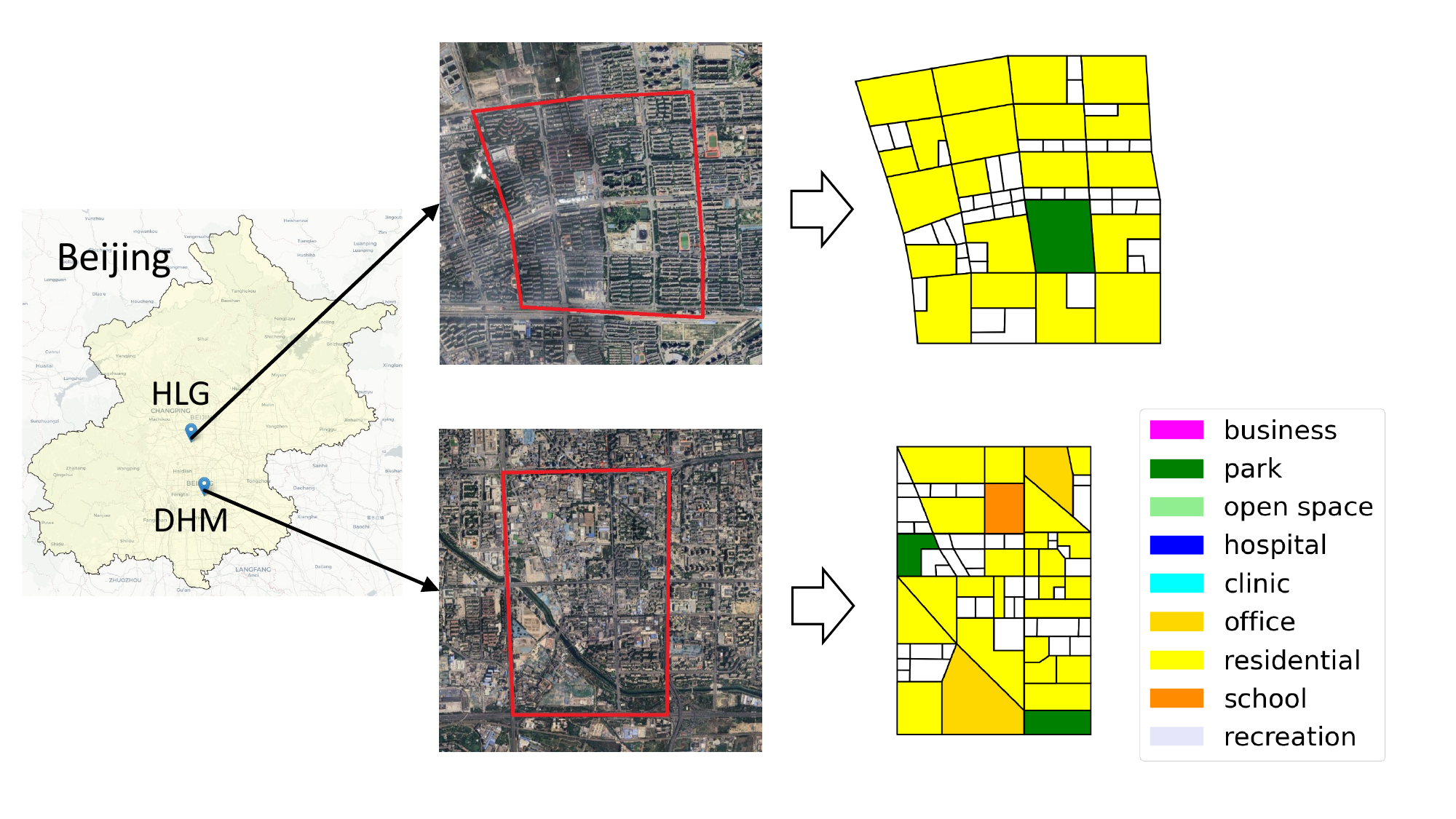}
    \caption{The satellite image and area division of two regions, Huilongguan (HLG) and Dahongmen (DHM).}
    \label{fig:dataset}
\end{figure}

We deploy our methods on two real-world regions, Huilongguan (HLG) and Dahongmen (DHM), as shown in Figure~\ref{fig:dataset}. 
\begin{itemize}
    \item \textbf{HLG}. Situated 33km north of Beijing's city center, HLG was originally a satellite town and has now integrated into the Beijing metropolis within the 6th Ring Road. Characterized by extensive high-rise residential complexes, HLG has evolved into one of the most densely populated regions in Beijing. However, the surrounding infrastructure is inadequate, and limited employment opportunities necessitate many residents to commute for remote work, making HLG a typical commuter town. It would be beneficial for HLG to be renovated to create a more livable environment.
    \item \textbf{DHM}. Located in south Beijing, DHM represents a different urban landscape. With a long history of residency, this region has experienced unplanned and haphazard development, resulting in a scattered urban layout that intertwines residential spaces with commerce, warehousing, logistics, and rental compounds. Apart from the disordered landscapes, some informal settlement areas lacking sufficient infrastructure have been preserved, rendering the area susceptible to flooding and other risks. Therefore, it is necessary to renovate DHM to achieve a more reasonable and efficient facility allocation.
\end{itemize}

\begin{table}[htbp!]
    \centering
    \caption{The basic information of two real-world datasets.}
    \resizebox{.95\linewidth}{!}{
\begin{tabular}{c|ccccc}
\hline
\textbf{Region} & \textbf{Total Area} & \textbf{\#Residents} & \textbf{\#Areas} & \textbf{\#Vacant Areas} & \textbf{\#Agents} \\ \hline
HLG             & 3.74 km$^2$         & 85,041               & 63               & 42                      & 1000              \\
DHM             & 5.17 km$^2$         & 73,130               & 70               & 42                      & 1000              \\ \hline
\end{tabular}
        }
    \label{tbl:dataset}
\end{table}

Apart from the necessity of renovation, the choice of regions also considers their diverse socio-economic profiles, demographic compositions, and complexities in urban planning. Our experiment aimed to capture a broad spectrum of participatory urban planning scenarios, fostering a comprehensive understanding of the LLM's effectiveness in diverse contexts.

\subsubsection{Implementation Details}
We partitioned HLG and DHM regions into areas based on roads and existing land-use conditions. Following common scenarios in actual urban redevelopment, we retain the residency and major green land and treat other areas as vacant spaces. Both HLG and DHM consist of 42 vacant areas to be planned. Here we restrict the available land use to 8 types~\citep{zheng2023spatial}, including school, hospital, clinic, business area, office area, recreation area, park, and open space.

To ensure that the plan is realistic and well-planned, we established some basic requirements. For both communities, a minimum number of areas are mandated to ensure adequate infrastructure coverage. For HLG, there need to be at least 6 schools, 2 hospitals, 4 clinics, 4 business areas, 6 office areas, 6 recreation areas, 2 parks, and 4 open spaces. For DHM, the requirements are 7 schools, 1 hospital, 4 clinics, 4 business areas, 2 office areas, 6 recreation areas, 2 parks, and 6 open spaces.

We collect the demographics data from the census yearbook~\citep{BMBS2022Beijing}, and generate 1000 agents for each region according to the methods in Section~\ref{sec:role_play}. It is noteworthy that the agent number surpasses 1\% of the actual population in both communities, aligning with the standard used in yearly censuses and exceeding the typical consideration in participatory planning.
Moreover, we incorporate marginalized groups with special backgrounds, including 10 elderly persons living alone, 10 families with a sick person, 50 parenting families, 50 families with children attending school, 50 Beijing drifters, and 50 office workers.

In the experiments, we use \textit{gpt-3.5-turbo-1106} for residents and \textit{gpt-4-vision-preview} for the planner since it needs the map of the region as input. During the renovation of each community, we let the residents discuss for 3 rounds, and 50 residents are selected to discuss in each round. To ensure the robustness, we set the temperature of LLMs as 0, and report the average results of five runs with different random seeds.

\subsubsection{Baselines}
To evaluate our framework, we selected six methods as baselines for comparison: the random method, the centralized method, the decentralized method, the Geometric Set Cover Algorithms (GSCA), the Deep Reinforcement Learning (DRL), and the result from human designers.

\begin{itemize}
    \item Random: We randomly assign each area with a land-use type.
    \item Centralized: For each land-use type, the probability of assigning a new area is inversely proportional to the distance to the community center, aiming for a compact arrangement~\citep{peter1997are}.
    \item Decentralized: For each land-use type, the probability of assigning a new area is proportional to the distance to the same type of areas~\citep{sakieh2015evaluating}, resulting in a decentralized distribution.
    \item GSCA: For each land-use, we solve the geometric-set-coverage-like problem by maximizing the coverage of the given land-use type~\citep{wei2016coverage}.
    \item Human Expert: We recruit professional human designers to accomplish the planning tasks.
    \item DRL~\citep{zheng2023spatial}: It uses the deep reinforcement learning method to maximize the metric \textit{Service} and \textit{Ecology}.
\end{itemize}

The implementation of baseline methods follows established procedures outlined in the literature~\citep{zheng2023spatial}. A team of 8 professional planners from the UK and China, with a minimum of 3 years of experience in urban planning, was recruited for the study. The results generated by human designers were transformed in ArcGIS format, enabling the subsequent computation of relevant metrics.

\begin{table*}[htbp!]
    \centering
    \caption{Performance comparison with baselines on two datasets. The best results are presented in bold, and the second-best results are underlined.}
    \resizebox{.8\linewidth}{!}{
\begin{tabular}{c|cccc|cccc}
\hline
                       & \multicolumn{4}{c|}{\textbf{HLG}}                                                & \multicolumn{4}{c}{\textbf{DHM}}                                                 \\
\textbf{Model}         & \textbf{Service} & \textbf{Ecology} & \textbf{Satisfaction} & \textbf{Inclusion} & \textbf{Service} & \textbf{Ecology} & \textbf{Satisfaction} & \textbf{Inclusion} \\ \hline
\textbf{Random}        & 0.491            & 0.505            & {\ul 0.708}           & 0.698              & 0.690            & 0.664            & {\ul 0.689}           & 0.698              \\
\textbf{Centralized}   & 0.654            & 0.364            & 0.578                 & 0.560              & 0.562            & 0.393            & 0.516                 & 0.538              \\
\textbf{Decentralized} & 0.709            & 0.455            & 0.678                 & 0.691              & {\ul 0.743}      & 0.518            & 0.687                 & {\ul 0.706}        \\
\textbf{GSCA}          & 0.682            & 0.439            & 0.653                 & 0.657              & 0.584            & 0.464            & 0.587                 & 0.621              \\
\textbf{Human Expert}  & {\ul 0.756}      & 0.468            & 0.650                 & 0.475              & 0.717            & 0.527            & 0.631                 & 0.544              \\
\textbf{DRL}           & \textbf{0.773}   & \textbf{0.747}   & {\ul 0.708}           & {\ul 0.716}        & 0.671            & \textbf{0.880}   & 0.566                 & 0.605              \\ \hline
\textbf{Ours}          & {\ul 0.756}      & {\ul 0.713}      & \textbf{0.787}        & \textbf{0.773}     & \textbf{0.760}   & {\ul 0.724}      & \textbf{0.778}        & \textbf{0.790}     \\
\textbf{Improvement}   & -2.2\%           & -4.6\%           & 11.2\%                & 8.0\%              & 2.3\%            & -17.7\%          & 12.9\%                & 11.9\%             \\ \hline
\end{tabular}
        }
    \label{tbl:result}
\end{table*}

\subsubsection{Evaluation Metrics}
\label{sec:metrics}
To evaluate the performance of our framework, we employed two categories of metrics: need-agnostic and need-aware. Need-agnostic metrics are aggregated indicators providing an overview of the entire community, focusing on service and ecology without considering the customized needs of individuals. 
\begin{itemize}
    \item \textbf{Service}~\citep{zheng2023spatial}. This metric measures the layout efficiency of public services, which is calculated as the average proportion of basic services (including education, medical care, working, shopping, and entertainment) that can be accessed within 500m of residents' homes, ranging from 0 to 1. Specifically, we first calculate the minimum distance $d(m, j)$ for the agent $m$ to access the $j$th type of areas $P^j$ as:
    \begin{equation}
        d(m, j) = \min \{ \mathit{EucDis}{}(L_m, P_1^j), \dots, \mathit{EucDis}(L_m, P_{k_j}^j) \}, 
    \end{equation}
    where $L_m$ represents the home location of agent $m$, and the $k_j$ is the total number of areas of type $j$. Subsequently, the \textit{Service} metric is defined as:
    \begin{equation}
        Service = \frac{1}{n_m} \sum_{m=1}^{n_m} \frac{1}{n_j} \sum_{j=1}^{n_j} \mathbbm{1}[d(m,j)<500],
    \end{equation}
    where $n_m$ denotes the number of agents and $n_j$ is the number of land use types.
    \item \textbf{Ecology}~\citep{zheng2023spatial}. Parks and open spaces are important to the health of residents. Therefore, we use this metric to measure the proportion of residents covered by the serving range of parks and open spaces. Specifically, we define the Ecology Service Range (ESR) as the union of buffers extending 300m from each park and open space $P_k^{park}$, represented as:
    \begin{equation}
        ESR = \mathit{Union} \{ \mathit{Buffer}(P_1^{park}, 300), \dots, \mathit{Buffer}(P_{k}^{park}, 300) \}.
    \end{equation}
    Subsequently, the \textit{Ecology} metric is defined as:
    \begin{equation}
        Ecology = \frac{1}{n_m}\sum_{m=1}^{n_m}\mathbbm{1}[L_i \in ESR],
    \end{equation}
    
\end{itemize}

The two need-agnostic metrics effectively encapsulate the concept of a 15-minute life circle~\citep{smartcities4010006}, ensuring that basic community services are reachable within 15 minutes by walking or cycling. 
However, these metrics do not consider the various needs of residents with different profiles. Therefore, we further introduce two need-aware metrics \textit{Satisfaction} and \textit{Inclusion}.

\begin{itemize}
    \item \textbf{Satisfaction}. At the forefront of participatory planning, the paramount consideration lies in meticulously examining residents' feedback and opinions. As mentioned before, each resident agent $m$ can report a set of land-use types comprising 3-5 deemed most urgently needed, denoted as $J_m$. 
    Therefore, we propose the \textit{Satisfaction} metric to quantify the extent to which each agent's needs are fulfilled, ranging from 0 to 1. Specifically, we define the satisfaction level for agent $m$ as:
    \begin{equation}
        S_m = \frac{1}{n_j} \sum_{j=1}^{n_j} \mathbbm{1}[d(m,j)<500], \quad j \in J_m ,
    \end{equation}
    where $d(m,j)$ represents the minimum distance for the $m$th agent to access the $j$th type of areas. The overall satisfaction metric is then calculated as:
    \begin{equation}
        Satisfaction = \frac{1}{n_m} \sum_{m=1}^{n_m} S_m,
    \end{equation}
    with $n_m$ denoting the total number of agents. 
    
    \item \textbf{Inclusion}. In parallel, we introduce an Inclusion metric to safeguard the interests of marginalized groups $V$ in a similar way:
    \begin{equation}
        S_v = \frac{1}{n_j} \sum_{j=1}^{n_j} \mathbbm{1}[d(v,j)<500], \quad j \in J_v, \quad v \in V, \\
    \end{equation}
    \begin{equation}
        Inclusion = \frac{1}{n_v} \sum_{v=1}^{n_v} S_v.        
    \end{equation}
    The \textit{Inclusion} measures whether the planning process adequately addresses the requirements of marginalized groups, contributing to a more equitable and inclusive urban development strategy. 
\end{itemize}

We combine the need-agnostic \textit{Service} and \textit{Ecology}, and need-aware \textit{Satisfaction} and \textit{Inclusion} to measure the performance of the intention draft, aiming to be in line with the core aspirations of participatory planning and taking care of the satisfaction of all stakeholders.

\subsection{Overall Performance}

The comparison of our methods and baselines is shown in Table~\ref{tbl:result}, from which we have the following observations.

First, our method achieves the best performance in need-aware metrics on both datasets, i.e., \textit{Satisfaction} and \textit{Inclusion}, outperforming baselines by a significant margin. For example, on the HLG dataset, our method's \textit{Satisfaction} score is 0.787, indicating that residents can access 78.7\% of the facilities they need within 500m of their homes on average, surpassing baselines by over 11.2\%. Moreover, the \textit{Inclusion} score 0.773 indicates that 77.3\% of the marginalized group's needs are satisfied. Such results demonstrate that our methods can effectively consider the diverse needs of residents by allowing residents to discuss and planner to modify the plan according to their opinions, which is exactly what "participatory" means.

Second, in terms of need-agnostic metrics, our method generally achieves the second-best performance, with only a slight gap compared to DRL, and outperforming human experts. Note that the DRL method is optimized for higher \textit{Service} and \textit{Ecology} scores. However, DRL fails to consider residents' specific needs, resulting in a much lower \textit{Satisfaction} and \textit{Inclusion} score.

Third, among baselines, human experts can achieve rather higher \textit{Service} scores on both datasets, indicating that human experts are good at planning the spatial layout of different land uses. The decentralized method also shows good performance in terms of \textit{Service} because it tends to place areas with the same land use far from each other, leading to high coverage and efficiency. However, all these methods do not incorporate residents. In comparison, the simulated residents in our methods focus on their own neighborhoods and propose suggestions to improve land-use distribution from a local view, which collaboratively achieves a better plan through discussion and aggregation.

In summary, the multi-agent collaboration design not only enables our method to effectively take the diverse needs of residents into account, but also brings our plan competitive public services accessibility and ecology score.

\subsection{Ablation Study}

\begin{figure}[htbp!]
\centering
\hspace{-3mm}
    \subfigure[HLG]{
    {\label{subfig:abl_hlg}}
    \includegraphics[width=.48\linewidth]{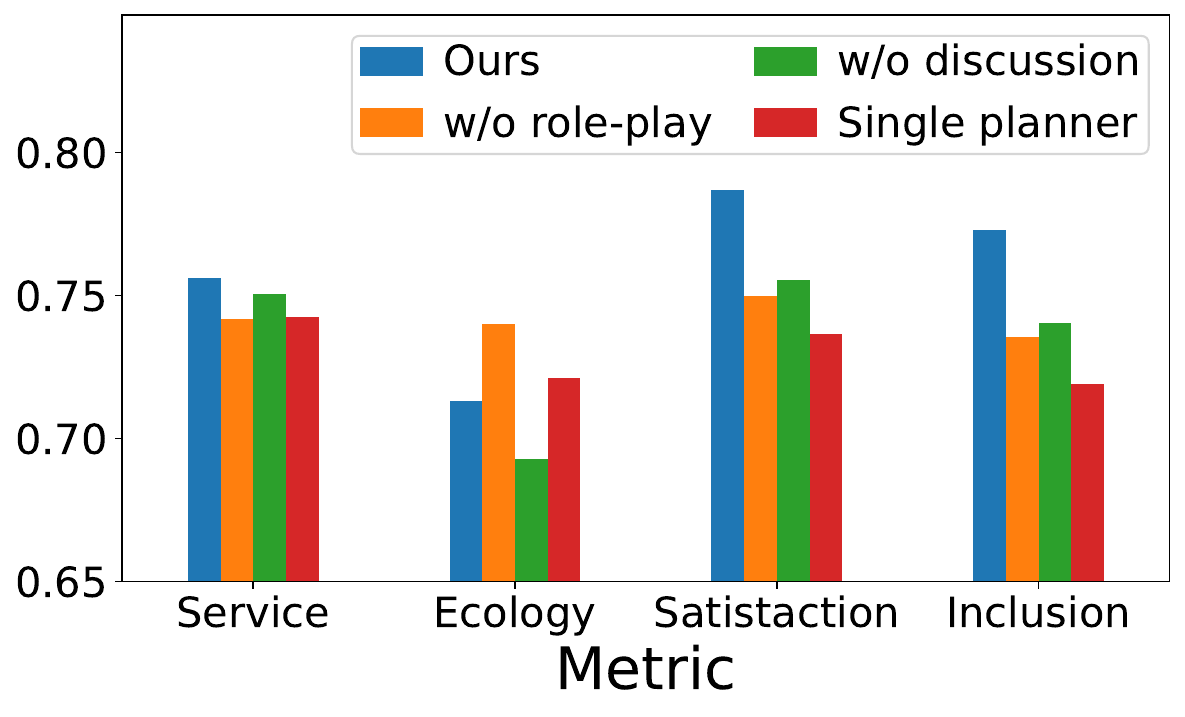}
    }
    \subfigure[DHM]{
    {\label{subfig:abl_dhm}}
    \includegraphics[width=.48\linewidth]{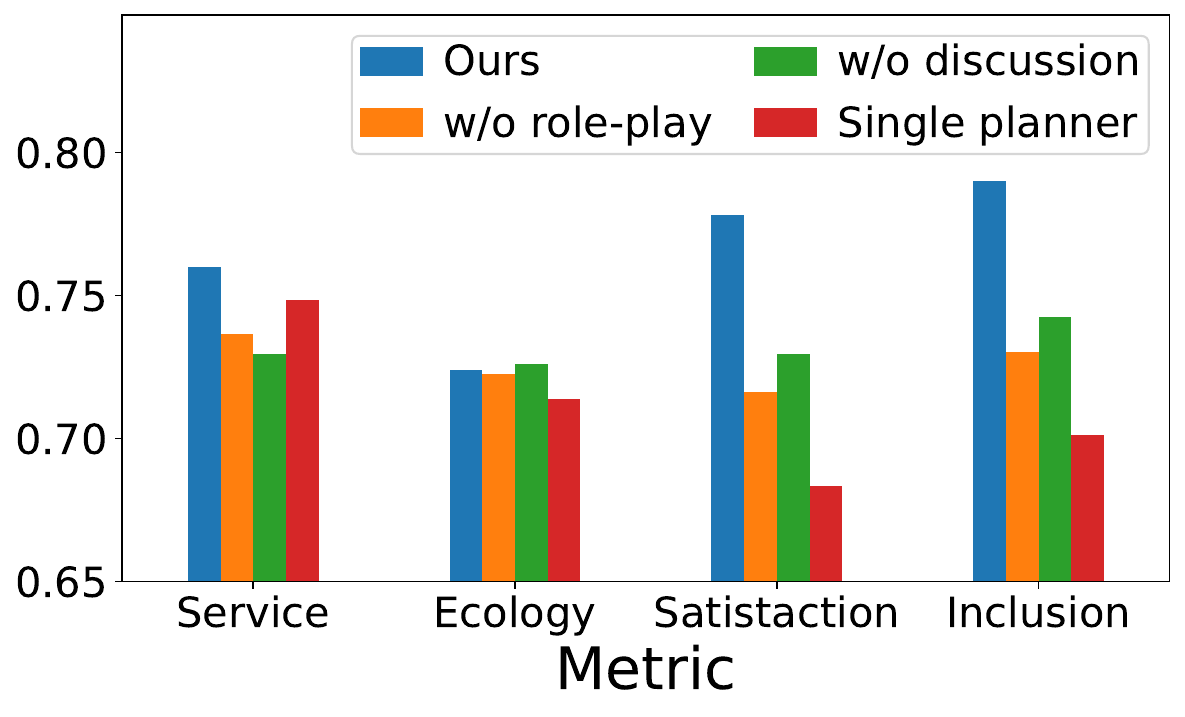}
    }
\hspace{-3mm}
\caption{Performance comparison of our methods, our method without role-play, our method without discussion, and single planner.
}\label{fig:ablation}
\end{figure}

\begin{figure*}[htbp!]
\centering
    \subfigure[]{
    {\label{subfig:plan_change_hlg}}
    \includegraphics[width=.6\linewidth]{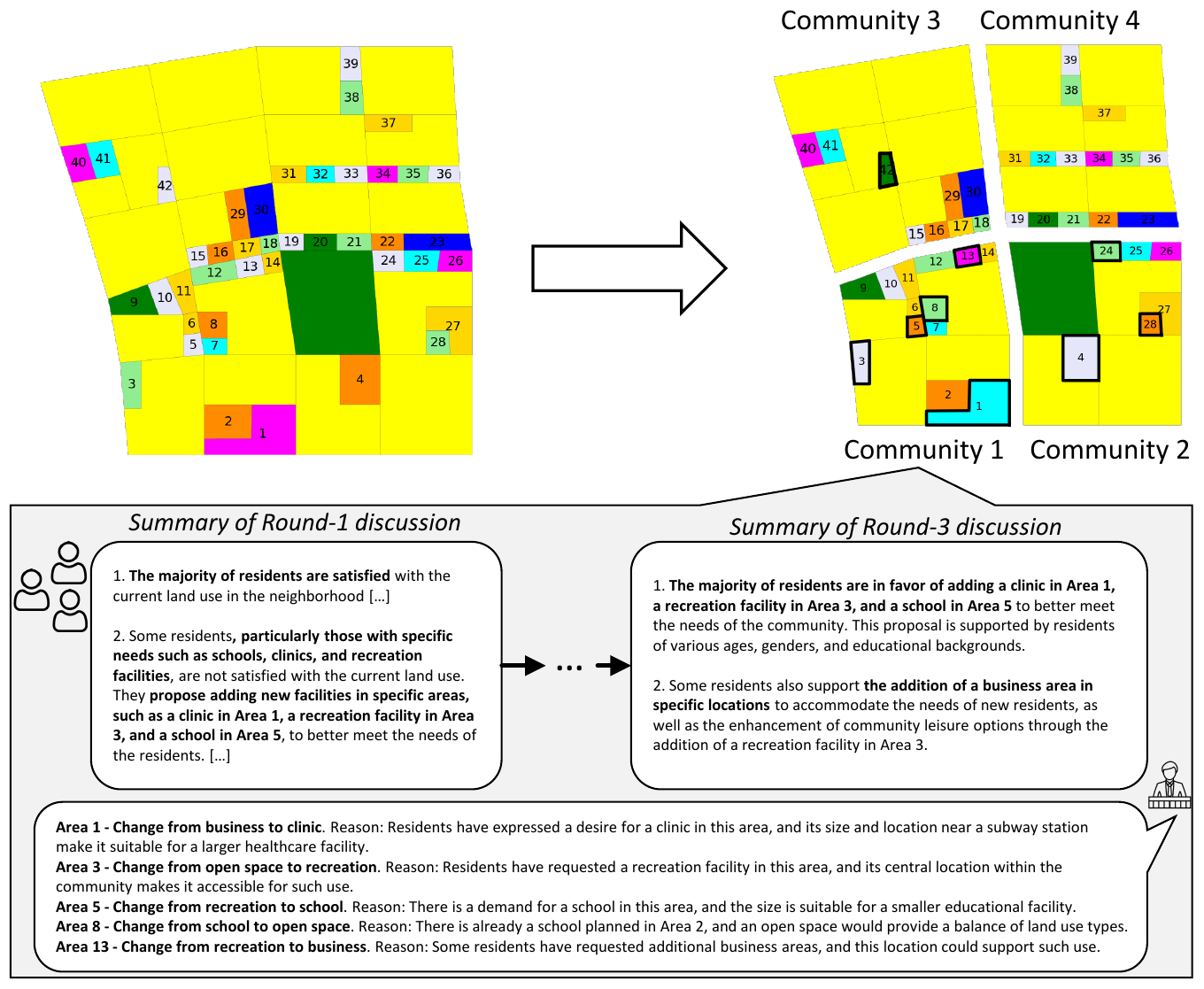}
    }
    \subfigure[]{
    {\label{subfig:metrics_change_hlg}}
    \includegraphics[width=.35\linewidth]{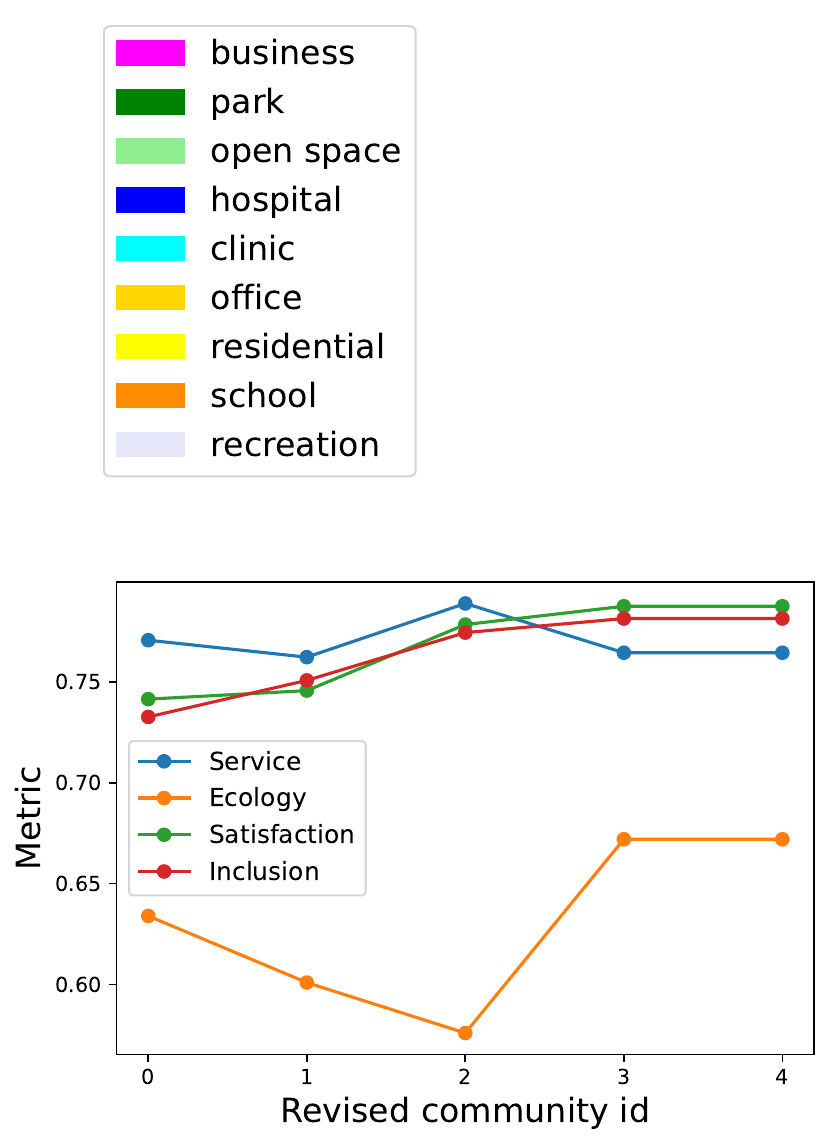}
    }
\caption{(a) The discussion of residents and revision of plan made by the planner of community 1 in HLG. Changes in land-use are marked in bold solid lines in the map. (b) The metrics after the revision of each community. 0 indicates the initial plan, and 4 indicates the final plan after revising 4 communities. The metrics are the average of 5 runs.}
\label{fig:plan_change}
\end{figure*}

We conduct an ablation study to evaluate the validity of role-play and discussion design. For role-play, we remove the profiles of all residents and replace them with the same prompt "You are a resident living in a region in the city", so residents cannot respond according to their profiles in the discussion (denoted as "w/o role-play"). Note that we still use the original needs of land use to calculate \textit{Satisfaction} and \textit{Inclusion}. To evaluate the effectiveness of discussion, we do not allow the residents to discuss with each other. Instead, the participants in the discussion only report their own opinions about the plan, which are then summarized and sent to the planner for plan revision (denoted as "w/o discussion"). Moreover, we also compare with the initial plan before discussion, i.e., the plan proposed by the planner alone (denoted as "Single planner").

The results are presented in Figure~\ref{fig:ablation}, all of which are the average of five runs. We can observe that the performance on almost all metrics drops after removing either part, demonstrating the effectiveness of role-play and discussion.
Specifically, the need-aware metrics, i.e., \textit{Satisfaction} and \textit{Inclusion}, drop 4.1\%-6.3\% after removing discussion, suggesting that simply asking and aggregating each resident's opinion separately is sub-optimal. On the contrary, the discussion allows residents to communicate with each other, which can better balance their different opinions and diverse needs.
Moreover, the lack of role-play also reduces the need-aware metrics by 4.7\%-8.0\%, which indicates the profiles of residents can effectively reflect their diverse needs.
Furthermore, it is not surprising that single planner achieves the lowest \textit{Satisfaction} and \textit{Inclusion}, as the planner does not consider residents at all. Such results demonstrate the importance of residents' participation in the planning process.
We also note that the decreases in need-agnostic metrics, \textit{Service} and \textit{Ecology}, are rather small, and the \textit{Ecology} even increases a little on the HLG dataset after ablation. This is probably because of the potential trade-off between metrics, as our methods tend to increase facilities to meet residents' needs, which may result in less green space.



\subsection{Analysis of Discussion}

To better understand the residents' discussion and planner revision process, we take HLG as an example and visualize the changes in lane use after revising each community in Figure~\ref{subfig:plan_change_hlg}. For example, the land-use types of Area 1, 3, 5, 8, and 13 are changed in Community 1, and Area 42 is changed to a park from a recreational area after the revision of Community 3. The residents of Community 4 seem to be satisfied with the initial plan as no areas are changed.

We also present the key parts of the dialogues generated during the revision of Community 1 below.
It can be observed that in the first round of discussion, the majority of residents are already satisfied with the plan, indicating that the initial plan proposed by the planner is reasonable to some degree. However, there are still some residents with special needs who propose adding new facilities in Area 1, 3, and 5. After several rounds of discussion, many other residents received such opinions and expressed their support. Therefore, the planner revised these areas according to the suggestions. Through such a process, the viewpoints of the minority can receive thorough discussion and consideration, which results in the improvement of the overall satisfaction of the community.
In summary, the planner generates a reasonably sound initial plan, and participatory discussions refine and enhance this outcome. The community-level discussion enables resident agents to concentrate more effectively on areas directly related to their interests and guarantees substantive inclusion of concerns in the planning process.

Moreover, recall that the four communities are revised sequentially, and we present the evaluation metrics after revising each community in Figure~\ref{subfig:metrics_change_hlg}. It can be observed that, with the gradual progression of modifications, need-aware metrics show a noticeable enhancement, while need-agnostic metrics fluctuate. This trend is unsurprising since participatory adjustments primarily consider individual needs. Simultaneously, we observe that these refinements have a minimal impact on need-agnostic metrics, suggesting rational adjustments on the initial planning can significantly improve \textit{Satisfaction} and \textit{Inclusion} without undermining facilities coverage. This further underscores the importance of the participatory method.

\subsection{Hyperparameter Study}

\begin{figure}[htbp!]
\centering
\hspace{-3mm}
    \subfigure[HLG]{
    {\label{subfig:nround_hlg}}
    \includegraphics[width=.48\linewidth]{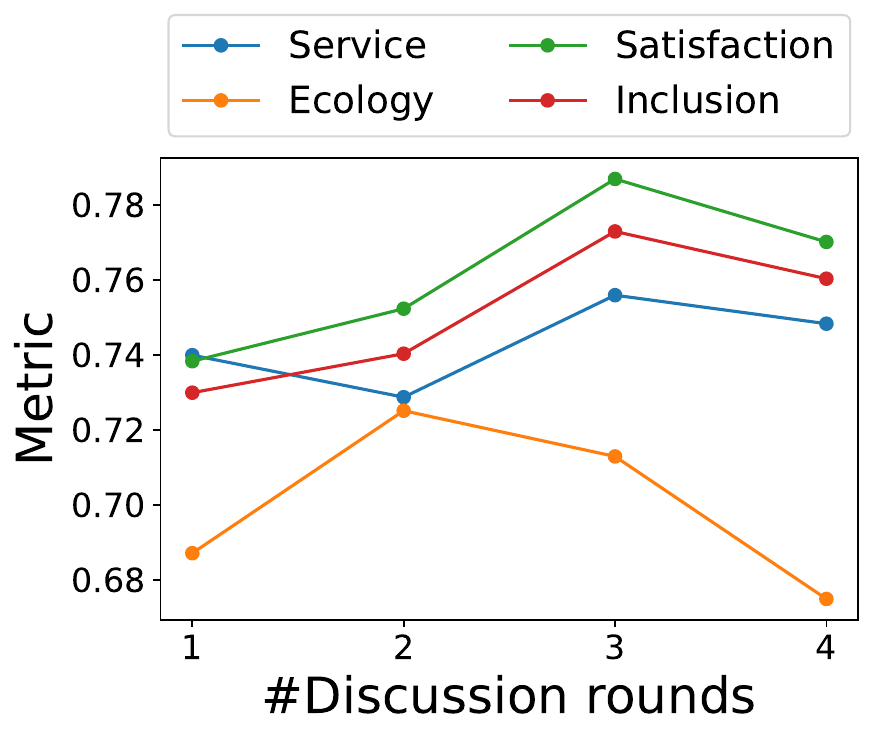}
    }
    \subfigure[DHM]{
    {\label{subfig:nround_dhm}}
    \includegraphics[width=.48\linewidth]{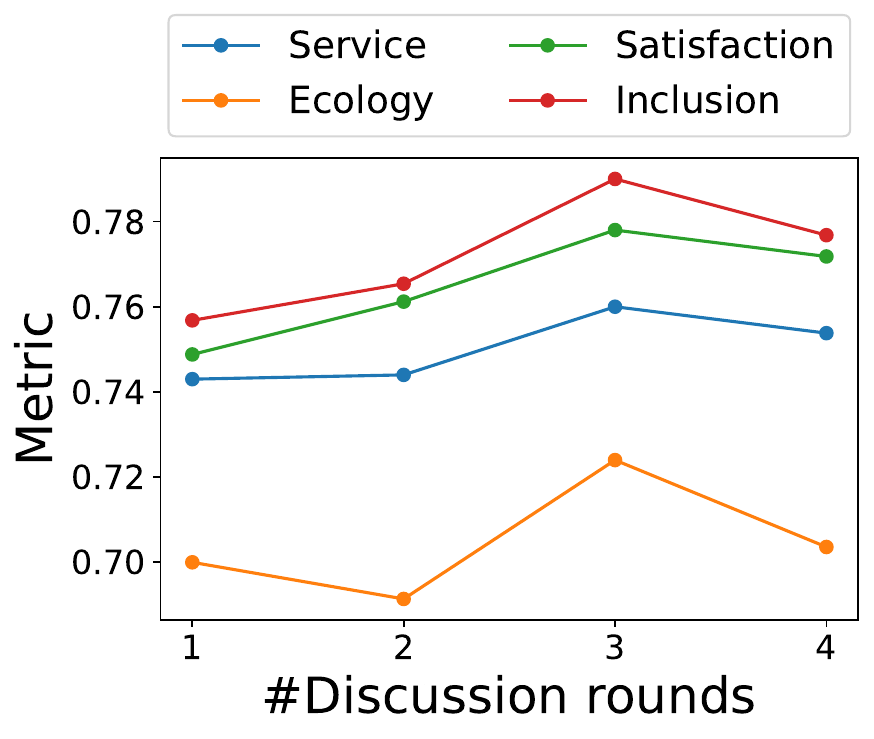}
    }
\hspace{-3mm}
\caption{Effect of the number of discussion rounds on the performance. The metrics are the average of 5 runs.
}\label{fig:nround}
\end{figure}

An important hyperparameter in our method is the number of rounds $N$ in the fishbowl discussion (Algorithm~\ref{alg:discussion}). To investigate the effect of $N$ on the outcome, we conduct experiments with $N=1,2,3,4$ and present the results in Figure~\ref{fig:nround}. As we can see, when $N$ increases from 1 to 3, there is a noticeable improvement in need-aware metrics, \textit{Satisfaction} and \textit{Inclusion}. For example, on HLG dataset, the $Satisfaction$ increases 6.6\% from 0.738 to 0.787, and $Inclusion$ increases 5.9\%. Such results demonstrate that multi-round discussion can better balance residents' diverse opinions and carry out land-use plan that meets their needs more effectively. Moreover, these metrics do not further go up when $N=4$, which is probably because overlong discussion may lead to stagnation or even deterioration of performance~\citep{chan2023chateval,liang2023encouraging,du2023improving}.
Moreover, We observe that the \textit{Service} score shows a similar trend with \textit{Satisfaction} and \textit{Inclusion}, while the \textit{Ecology} score fluctuates, which may also result from the trade-off between facilities and green space.

\section{Conclusion and Discussion}
\label{sec:conclusion}


In this paper, we leverage LLMs to simulate the entire process of participatory urban planning. Specifically, we craft LLM agents to simulate the planner and residents with different profiles through specific prompt design. Then, we propose a framework where residents discuss the land-use plan made by the planner, and the planner revises the plan accordingly. Moreover, to enable more efficient discussion among a large number of residents, we adopt a fishbowl discussion mechanism where part of the residents discuss and the rest listen for several rounds. Our method is deployed on two real-world scenarios in Beijing, and the results show that our method can effectively satisfy residents' diverse needs, and is competitive to state-of-the-art reinforcement learning methods in terms of service accessibility and ecology metrics.

This study is only an initial effort to apply LLMs to participatory urban planning. We must acknowledge the numerous simplifications regarding planning elements involved throughout the process. Factors such as ownership, development costs, and higher-level planning were omitted in this study, potentially leading to a deviation from practical planning scenarios. These shortcomings can hopefully be addressed through new prompts without compromising the effectiveness and interpretability of our framework.
It is also noteworthy that our framework largely relies on manually designed prompts, especially for the planner. To enable the planner agent to understand the planning map, we need to describe the location and neighborhood of each area, limiting the generalization ability of our method.


As for future work, a promising direction is integrating our framework with human experts to construct a human-AI collaborative workflow~\citep{zheng2023spatial}. For example, we can replace the planner or some residents with humans for better realness and performance. Moreover, the multi-agent collaboration design can hopefully be applied to other tasks that involve a large number of agents~\citep{qi2024civrealm}.

\bibliographystyle{ACM-Reference-Format}
\bibliography{reference}


\begin{thebibliography}{39}


\ifx \showCODEN    \undefined \def \showCODEN     #1{\unskip}     \fi
\ifx \showDOI      \undefined \def \showDOI       #1{#1}\fi
\ifx \showISBNx    \undefined \def \showISBNx     #1{\unskip}     \fi
\ifx \showISBNxiii \undefined \def \showISBNxiii  #1{\unskip}     \fi
\ifx \showISSN     \undefined \def \showISSN      #1{\unskip}     \fi
\ifx \showLCCN     \undefined \def \showLCCN      #1{\unskip}     \fi
\ifx \shownote     \undefined \def \shownote      #1{#1}          \fi
\ifx \showarticletitle \undefined \def \showarticletitle #1{#1}   \fi
\ifx \showURL      \undefined \def \showURL       {\relax}        \fi
\providecommand\bibfield[2]{#2}
\providecommand\bibinfo[2]{#2}
\providecommand\natexlab[1]{#1}
\providecommand\showeprint[2][]{arXiv:#2}

\bibitem[Abas et~al\mbox{.}(2023)]%
        {abas2023systematic}
\bibfield{author}{\bibinfo{person}{Azlan Abas}, \bibinfo{person}{Kadir Arifin}, \bibinfo{person}{Mohd Azhar~Mohamed Ali}, {and} \bibinfo{person}{Muhammad Khairil}.} \bibinfo{year}{2023}\natexlab{}.
\newblock \showarticletitle{A systematic literature review on public participation in decision-making for local authority planning: A decade of progress and challenges}.
\newblock \bibinfo{journal}{\emph{Environmental Development}}  \bibinfo{volume}{46} (\bibinfo{year}{2023}), \bibinfo{pages}{100853}.
\newblock
\showISSN{2211-4645}
\urldef\tempurl%
\url{https://doi.org/10.1016/j.envdev.2023.100853}
\showDOI{\tempurl}


\bibitem[{Ahmadi Oloonabadi} and Baran(2023)]%
        {ahmadi2023augmented}
\bibfield{author}{\bibinfo{person}{Saeed {Ahmadi Oloonabadi}} {and} \bibinfo{person}{Perver Baran}.} \bibinfo{year}{2023}\natexlab{}.
\newblock \showarticletitle{Augmented reality participatory platform: A novel digital participatory planning tool to engage under-resourced communities in improving neighborhood walkability}.
\newblock \bibinfo{journal}{\emph{Cities}}  \bibinfo{volume}{141} (\bibinfo{year}{2023}), \bibinfo{pages}{104441}.
\newblock
\showISSN{0264-2751}
\urldef\tempurl%
\url{https://doi.org/10.1016/j.cities.2023.104441}
\showDOI{\tempurl}


\bibitem[Anonymous(2024)]%
        {anonymous2024chainofexperts}
\bibfield{author}{\bibinfo{person}{Anonymous}.} \bibinfo{year}{2024}\natexlab{}.
\newblock \showarticletitle{Chain-of-Experts: When {LLM}s Meet Complex Operations Research Problems}. In \bibinfo{booktitle}{\emph{The Twelfth International Conference on Learning Representations}}.
\newblock
\urldef\tempurl%
\url{https://openreview.net/forum?id=HobyL1B9CZ}
\showURL{%
\tempurl}


\bibitem[Arnstein(1969)]%
        {arnstein1969ladder}
\bibfield{author}{\bibinfo{person}{Sherry~R. Arnstein}.} \bibinfo{year}{1969}\natexlab{}.
\newblock \showarticletitle{A Ladder Of Citizen Participation}.
\newblock \bibinfo{journal}{\emph{Journal of the American Institute of Planners}} \bibinfo{volume}{35}, \bibinfo{number}{4} (\bibinfo{year}{1969}), \bibinfo{pages}{216--224}.
\newblock
\urldef\tempurl%
\url{https://doi.org/10.1080/01944366908977225}
\showDOI{\tempurl}


\bibitem[{Beijing Municipal Bureau of Statistics} and {Survey Office of the National Bureau of Statistics in Bejing}(2022)]%
        {BMBS2022Beijing}
\bibfield{author}{\bibinfo{person}{{Beijing Municipal Bureau of Statistics}} {and} \bibinfo{person}{{Survey Office of the National Bureau of Statistics in Bejing}}.} \bibinfo{year}{2022}\natexlab{}.
\newblock \bibinfo{booktitle}{\emph{Beijing Population Census Yearbook: 2020}}.
\newblock \bibinfo{publisher}{China Statistics Press}.
\newblock
\showISBNx{9787503798054}
\urldef\tempurl%
\url{https://nj.tjj.beijing.gov.cn/tjnj/rkpc-2020/indexch.htm}
\showURL{%
\tempurl}


\bibitem[Chan et~al\mbox{.}(2023)]%
        {chan2023chateval}
\bibfield{author}{\bibinfo{person}{Chi-Min Chan}, \bibinfo{person}{Weize Chen}, \bibinfo{person}{Yusheng Su}, \bibinfo{person}{Jianxuan Yu}, \bibinfo{person}{Wei Xue}, \bibinfo{person}{Shanghang Zhang}, \bibinfo{person}{Jie Fu}, {and} \bibinfo{person}{Zhiyuan Liu}.} \bibinfo{year}{2023}\natexlab{}.
\newblock \bibinfo{title}{ChatEval: Towards Better LLM-based Evaluators through Multi-Agent Debate}.
\newblock
\newblock
\showeprint[arxiv]{2308.07201}~[cs.CL]


\bibitem[Chassin et~al\mbox{.}(2022)]%
        {chassin2022experiencing}
\bibfield{author}{\bibinfo{person}{Thibaud Chassin}, \bibinfo{person}{Jens Ingensand}, \bibinfo{person}{Sidonie Christophe}, {and} \bibinfo{person}{Guillaume Touya}.} \bibinfo{year}{2022}\natexlab{}.
\newblock \showarticletitle{Experiencing virtual geographic environment in urban 3D participatory e-planning: A user perspective}.
\newblock \bibinfo{journal}{\emph{Landscape and Urban Planning}}  \bibinfo{volume}{224} (\bibinfo{year}{2022}), \bibinfo{pages}{104432}.
\newblock
\showISSN{0169-2046}
\urldef\tempurl%
\url{https://doi.org/10.1016/j.landurbplan.2022.104432}
\showDOI{\tempurl}


\bibitem[Chen et~al\mbox{.}(2023)]%
        {chen2023agentverse}
\bibfield{author}{\bibinfo{person}{Weize Chen}, \bibinfo{person}{Yusheng Su}, \bibinfo{person}{Jingwei Zuo}, \bibinfo{person}{Cheng Yang}, \bibinfo{person}{Chenfei Yuan}, \bibinfo{person}{Chi-Min Chan}, \bibinfo{person}{Heyang Yu}, \bibinfo{person}{Yaxi Lu}, \bibinfo{person}{Yi-Hsin Hung}, \bibinfo{person}{Chen Qian}, \bibinfo{person}{Yujia Qin}, \bibinfo{person}{Xin Cong}, \bibinfo{person}{Ruobing Xie}, \bibinfo{person}{Zhiyuan Liu}, \bibinfo{person}{Maosong Sun}, {and} \bibinfo{person}{Jie Zhou}.} \bibinfo{year}{2023}\natexlab{}.
\newblock \bibinfo{title}{AgentVerse: Facilitating Multi-Agent Collaboration and Exploring Emergent Behaviors}.
\newblock
\newblock
\showeprint[arxiv]{2308.10848}~[cs.CL]


\bibitem[Du et~al\mbox{.}(2023b)]%
        {du2023artificial}
\bibfield{author}{\bibinfo{person}{Jiaxin Du}, \bibinfo{person}{Xinyue Ye}, \bibinfo{person}{Piotr Jankowski}, \bibinfo{person}{Thomas~W. Sanchez}, {and} \bibinfo{person}{Gengchen Mai}.} \bibinfo{year}{2023}\natexlab{b}.
\newblock \showarticletitle{Artificial intelligence enabled participatory planning: a review}.
\newblock \bibinfo{journal}{\emph{International Journal of Urban Sciences}} \bibinfo{volume}{0}, \bibinfo{number}{0} (\bibinfo{year}{2023}), \bibinfo{pages}{1--28}.
\newblock
\urldef\tempurl%
\url{https://doi.org/10.1080/12265934.2023.2262427}
\showDOI{\tempurl}


\bibitem[Du et~al\mbox{.}(2023a)]%
        {du2023improving}
\bibfield{author}{\bibinfo{person}{Yilun Du}, \bibinfo{person}{Shuang Li}, \bibinfo{person}{Antonio Torralba}, \bibinfo{person}{Joshua~B Tenenbaum}, {and} \bibinfo{person}{Igor Mordatch}.} \bibinfo{year}{2023}\natexlab{a}.
\newblock \showarticletitle{Improving Factuality and Reasoning in Language Models through Multiagent Debate}.
\newblock \bibinfo{journal}{\emph{arXiv preprint arXiv:2305.14325}} (\bibinfo{year}{2023}).
\newblock


\bibitem[Erik~Eriksson and Syssner(2022)]%
        {eriksson2022opening}
\bibfield{author}{\bibinfo{person}{Amira~Fredriksson Erik~Eriksson} {and} \bibinfo{person}{Josefina Syssner}.} \bibinfo{year}{2022}\natexlab{}.
\newblock \showarticletitle{Opening the black box of participatory planning: a study of how planners handle citizens’ input}.
\newblock \bibinfo{journal}{\emph{European Planning Studies}} \bibinfo{volume}{30}, \bibinfo{number}{6} (\bibinfo{year}{2022}), \bibinfo{pages}{994--1012}.
\newblock
\urldef\tempurl%
\url{https://doi.org/10.1080/09654313.2021.1895974}
\showDOI{\tempurl}


\bibitem[Forester(1982)]%
        {forester1982planning}
\bibfield{author}{\bibinfo{person}{John Forester}.} \bibinfo{year}{1982}\natexlab{}.
\newblock \showarticletitle{Planning in the Face of Power}.
\newblock \bibinfo{journal}{\emph{Journal of the American Planning Association}} \bibinfo{volume}{48}, \bibinfo{number}{1} (\bibinfo{year}{1982}), \bibinfo{pages}{67--80}.
\newblock
\urldef\tempurl%
\url{https://doi.org/10.1080/01944368208976167}
\showDOI{\tempurl}


\bibitem[Forester(1999)]%
        {forester1999deliberative}
\bibfield{author}{\bibinfo{person}{J. Forester}.} \bibinfo{year}{1999}\natexlab{}.
\newblock \bibinfo{booktitle}{\emph{The Deliberative Practitioner: Encouraging Participatory Planning Processes}}.
\newblock \bibinfo{publisher}{MIT Press}.
\newblock
\showISBNx{9780262561228}
\showLCCN{99015445}
\urldef\tempurl%
\url{https://books.google.com/books?id=ywJXreTLoBcC}
\showURL{%
\tempurl}


\bibitem[Gordon and Richardson(1997)]%
        {peter1997are}
\bibfield{author}{\bibinfo{person}{Peter Gordon} {and} \bibinfo{person}{Harry~W. Richardson}.} \bibinfo{year}{1997}\natexlab{}.
\newblock \showarticletitle{Are Compact Cities a Desirable Planning Goal?}
\newblock \bibinfo{journal}{\emph{Journal of the American Planning Association}} \bibinfo{volume}{63}, \bibinfo{number}{1} (\bibinfo{year}{1997}), \bibinfo{pages}{95--106}.
\newblock
\urldef\tempurl%
\url{https://doi.org/10.1080/01944369708975727}
\showDOI{\tempurl}


\bibitem[Hong et~al\mbox{.}(2023)]%
        {hong2023metagpt}
\bibfield{author}{\bibinfo{person}{Sirui Hong}, \bibinfo{person}{Mingchen Zhuge}, \bibinfo{person}{Jonathan Chen}, \bibinfo{person}{Xiawu Zheng}, \bibinfo{person}{Yuheng Cheng}, \bibinfo{person}{Ceyao Zhang}, \bibinfo{person}{Jinlin Wang}, \bibinfo{person}{Zili Wang}, \bibinfo{person}{Steven Ka~Shing Yau}, \bibinfo{person}{Zijuan Lin}, \bibinfo{person}{Liyang Zhou}, \bibinfo{person}{Chenyu Ran}, \bibinfo{person}{Lingfeng Xiao}, \bibinfo{person}{Chenglin Wu}, {and} \bibinfo{person}{Jürgen Schmidhuber}.} \bibinfo{year}{2023}\natexlab{}.
\newblock \bibinfo{title}{MetaGPT: Meta Programming for A Multi-Agent Collaborative Framework}.
\newblock
\newblock
\showeprint[arxiv]{2308.00352}~[cs.AI]


\bibitem[Jiang et~al\mbox{.}(2018)]%
        {jiang2018demonstrator}
\bibfield{author}{\bibinfo{person}{Like Jiang}, \bibinfo{person}{Massimiliano Masullo}, \bibinfo{person}{Luigi Maffei}, \bibinfo{person}{Fanyu Meng}, {and} \bibinfo{person}{Michael Vorländer}.} \bibinfo{year}{2018}\natexlab{}.
\newblock \showarticletitle{A demonstrator tool of web-based virtual reality for participatory evaluation of urban sound environment}.
\newblock \bibinfo{journal}{\emph{Landscape and Urban Planning}}  \bibinfo{volume}{170} (\bibinfo{year}{2018}), \bibinfo{pages}{276--282}.
\newblock
\showISSN{0169-2046}
\urldef\tempurl%
\url{https://doi.org/10.1016/j.landurbplan.2017.09.007}
\showDOI{\tempurl}


\bibitem[Koenig et~al\mbox{.}(2020)]%
        {koenig2020inteegrating}
\bibfield{author}{\bibinfo{person}{Reinhard Koenig}, \bibinfo{person}{Yufan Miao}, \bibinfo{person}{Anna Aichinger}, \bibinfo{person}{Katja Knecht}, {and} \bibinfo{person}{Kateryna Konieva}.} \bibinfo{year}{2020}\natexlab{}.
\newblock \showarticletitle{Integrating urban analysis, generative design, and evolutionary optimization for solving urban design problems}.
\newblock \bibinfo{journal}{\emph{Environment and Planning B: Urban Analytics and City Science}} \bibinfo{volume}{47}, \bibinfo{number}{6} (\bibinfo{year}{2020}), \bibinfo{pages}{997--1013}.
\newblock
\urldef\tempurl%
\url{https://doi.org/10.1177/2399808319894986}
\showDOI{\tempurl}


\bibitem[Li et~al\mbox{.}(2023b)]%
        {li2023camel}
\bibfield{author}{\bibinfo{person}{Guohao Li}, \bibinfo{person}{Hasan Abed Al~Kader Hammoud}, \bibinfo{person}{Hani Itani}, \bibinfo{person}{Dmitrii Khizbullin}, {and} \bibinfo{person}{Bernard Ghanem}.} \bibinfo{year}{2023}\natexlab{b}.
\newblock \showarticletitle{{CAMEL}: Communicative Agents for ''Mind'' Exploration of Large Language Model Society}. In \bibinfo{booktitle}{\emph{Thirty-seventh Conference on Neural Information Processing Systems}}.
\newblock
\urldef\tempurl%
\url{https://openreview.net/forum?id=3IyL2XWDkG}
\showURL{%
\tempurl}


\bibitem[Li et~al\mbox{.}(2023a)]%
        {li-etal-2023-theory}
\bibfield{author}{\bibinfo{person}{Huao Li}, \bibinfo{person}{Yu Chong}, \bibinfo{person}{Simon Stepputtis}, \bibinfo{person}{Joseph Campbell}, \bibinfo{person}{Dana Hughes}, \bibinfo{person}{Charles Lewis}, {and} \bibinfo{person}{Katia Sycara}.} \bibinfo{year}{2023}\natexlab{a}.
\newblock \showarticletitle{Theory of Mind for Multi-Agent Collaboration via Large Language Models}. In \bibinfo{booktitle}{\emph{Proceedings of the 2023 Conference on Empirical Methods in Natural Language Processing}}, \bibfield{editor}{\bibinfo{person}{Houda Bouamor}, \bibinfo{person}{Juan Pino}, {and} \bibinfo{person}{Kalika Bali}} (Eds.). \bibinfo{publisher}{Association for Computational Linguistics}, \bibinfo{address}{Singapore}, \bibinfo{pages}{180--192}.
\newblock
\urldef\tempurl%
\url{https://doi.org/10.18653/v1/2023.emnlp-main.13}
\showDOI{\tempurl}


\bibitem[Li et~al\mbox{.}(2020)]%
        {li2020collaborative}
\bibfield{author}{\bibinfo{person}{Xun Li}, \bibinfo{person}{Fan Zhang}, \bibinfo{person}{Eddie~Chi man Hui}, {and} \bibinfo{person}{Wei Lang}.} \bibinfo{year}{2020}\natexlab{}.
\newblock \showarticletitle{Collaborative workshop and community participation: A new approach to urban regeneration in China}.
\newblock \bibinfo{journal}{\emph{Cities}}  \bibinfo{volume}{102} (\bibinfo{year}{2020}), \bibinfo{pages}{102743}.
\newblock
\showISSN{0264-2751}
\urldef\tempurl%
\url{https://doi.org/10.1016/j.cities.2020.102743}
\showDOI{\tempurl}


\bibitem[Liang et~al\mbox{.}(2023)]%
        {liang2023encouraging}
\bibfield{author}{\bibinfo{person}{Tian Liang}, \bibinfo{person}{Zhiwei He}, \bibinfo{person}{Wenxiang Jiao}, \bibinfo{person}{Xing Wang}, \bibinfo{person}{Yan Wang}, \bibinfo{person}{Rui Wang}, \bibinfo{person}{Yujiu Yang}, \bibinfo{person}{Zhaopeng Tu}, {and} \bibinfo{person}{Shuming Shi}.} \bibinfo{year}{2023}\natexlab{}.
\newblock \showarticletitle{Encouraging Divergent Thinking in Large Language Models through Multi-Agent Debate}.
\newblock \bibinfo{journal}{\emph{arXiv preprint arXiv:2305.19118}} (\bibinfo{year}{2023}).
\newblock


\bibitem[Liu(2022)]%
        {liu2022participatory}
\bibfield{author}{\bibinfo{person}{Jiayan Liu}.} \bibinfo{year}{2022}\natexlab{}.
\newblock \bibinfo{booktitle}{\emph{Participatory Community Planning and Design Toolkit}}.
\newblock \bibinfo{publisher}{China Architecture \& Building Press}.
\newblock


\bibitem[Lock et~al\mbox{.}(2021)]%
        {lock2021towards}
\bibfield{author}{\bibinfo{person}{Oliver Lock}, \bibinfo{person}{Michael Bain}, {and} \bibinfo{person}{Christopher Pettit}.} \bibinfo{year}{2021}\natexlab{}.
\newblock \showarticletitle{Towards the collaborative development of machine learning techniques in planning support systems – a Sydney example}.
\newblock \bibinfo{journal}{\emph{Environment and Planning B: Urban Analytics and City Science}} \bibinfo{volume}{48}, \bibinfo{number}{3} (\bibinfo{year}{2021}), \bibinfo{pages}{484--502}.
\newblock
\urldef\tempurl%
\url{https://doi.org/10.1177/2399808320939974}
\showDOI{\tempurl}


\bibitem[Moreno et~al\mbox{.}(2021)]%
        {smartcities4010006}
\bibfield{author}{\bibinfo{person}{Carlos Moreno}, \bibinfo{person}{Zaheer Allam}, \bibinfo{person}{Didier Chabaud}, \bibinfo{person}{Catherine Gall}, {and} \bibinfo{person}{Florent Pratlong}.} \bibinfo{year}{2021}\natexlab{}.
\newblock \showarticletitle{Introducing the “15-Minute City”: Sustainability, Resilience and Place Identity in Future Post-Pandemic Cities}.
\newblock \bibinfo{journal}{\emph{Smart Cities}} \bibinfo{volume}{4}, \bibinfo{number}{1} (\bibinfo{year}{2021}), \bibinfo{pages}{93--111}.
\newblock
\showISSN{2624-6511}
\urldef\tempurl%
\url{https://doi.org/10.3390/smartcities4010006}
\showDOI{\tempurl}


\bibitem[Nasr-Azadani et~al\mbox{.}(2022)]%
        {nasrazadani2022rapid}
\bibfield{author}{\bibinfo{person}{Ellie Nasr-Azadani}, \bibinfo{person}{Denice Wardrop}, {and} \bibinfo{person}{Robert Brooks}.} \bibinfo{year}{2022}\natexlab{}.
\newblock \showarticletitle{Is the rapid development of visualization techniques enhancing the quality of public participation in natural resource policy and management? A systematic review}.
\newblock \bibinfo{journal}{\emph{Landscape and Urban Planning}}  \bibinfo{volume}{228} (\bibinfo{year}{2022}), \bibinfo{pages}{104586}.
\newblock
\showISSN{0169-2046}
\urldef\tempurl%
\url{https://doi.org/10.1016/j.landurbplan.2022.104586}
\showDOI{\tempurl}


\bibitem[Park et~al\mbox{.}(2023)]%
        {park2023generative}
\bibfield{author}{\bibinfo{person}{Joon~Sung Park}, \bibinfo{person}{Joseph O'Brien}, \bibinfo{person}{Carrie~Jun Cai}, \bibinfo{person}{Meredith~Ringel Morris}, \bibinfo{person}{Percy Liang}, {and} \bibinfo{person}{Michael~S. Bernstein}.} \bibinfo{year}{2023}\natexlab{}.
\newblock \showarticletitle{Generative Agents: Interactive Simulacra of Human Behavior}. In \bibinfo{booktitle}{\emph{Proceedings of the 36th Annual ACM Symposium on User Interface Software and Technology}} (<conf-loc>, <city>San Francisco</city>, <state>CA</state>, <country>USA</country>, </conf-loc>) \emph{(\bibinfo{series}{UIST '23})}. \bibinfo{publisher}{Association for Computing Machinery}, \bibinfo{address}{New York, NY, USA}, Article \bibinfo{articleno}{2}, \bibinfo{numpages}{22}~pages.
\newblock
\showISBNx{9798400701320}
\urldef\tempurl%
\url{https://doi.org/10.1145/3586183.3606763}
\showDOI{\tempurl}


\bibitem[Qi et~al\mbox{.}(2024)]%
        {qi2024civrealm}
\bibfield{author}{\bibinfo{person}{Siyuan Qi}, \bibinfo{person}{Shuo Chen}, \bibinfo{person}{Yexin Li}, \bibinfo{person}{Xiangyu Kong}, \bibinfo{person}{Junqi Wang}, \bibinfo{person}{Bangcheng Yang}, \bibinfo{person}{Pring Wong}, \bibinfo{person}{Yifan Zhong}, \bibinfo{person}{Xiaoyuan Zhang}, \bibinfo{person}{Zhaowei Zhang}, \bibinfo{person}{Nian Liu}, \bibinfo{person}{Wei Wang}, \bibinfo{person}{Yaodong Yang}, {and} \bibinfo{person}{Song-Chun Zhu}.} \bibinfo{year}{2024}\natexlab{}.
\newblock \bibinfo{title}{CivRealm: A Learning and Reasoning Odyssey in Civilization for Decision-Making Agents}.
\newblock
\newblock
\showeprint[arxiv]{2401.10568}~[cs.AI]


\bibitem[Qian et~al\mbox{.}(2023)]%
        {qian2023ai}
\bibfield{author}{\bibinfo{person}{Kejiang Qian}, \bibinfo{person}{Lingjun Mao}, \bibinfo{person}{Xin Liang}, \bibinfo{person}{Yimin Ding}, \bibinfo{person}{Jin Gao}, \bibinfo{person}{Xinran Wei}, \bibinfo{person}{Ziyi Guo}, {and} \bibinfo{person}{Jiajie Li}.} \bibinfo{year}{2023}\natexlab{}.
\newblock \bibinfo{title}{AI Agent as Urban Planner: Steering Stakeholder Dynamics in Urban Planning via Consensus-based Multi-Agent Reinforcement Learning}.
\newblock
\newblock
\showeprint[arxiv]{2310.16772}~[cs.AI]


\bibitem[Quan(2022)]%
        {quan2022urban}
\bibfield{author}{\bibinfo{person}{Steven~Jige Quan}.} \bibinfo{year}{2022}\natexlab{}.
\newblock \showarticletitle{Urban-GAN: An artificial intelligence-aided computation system for plural urban design}.
\newblock \bibinfo{journal}{\emph{Environment and Planning B: Urban Analytics and City Science}} \bibinfo{volume}{49}, \bibinfo{number}{9} (\bibinfo{year}{2022}), \bibinfo{pages}{2500--2515}.
\newblock
\urldef\tempurl%
\url{https://doi.org/10.1177/23998083221100550}
\showDOI{\tempurl}


\bibitem[Sakieh et~al\mbox{.}(2015)]%
        {sakieh2015evaluating}
\bibfield{author}{\bibinfo{person}{Yousef Sakieh}, \bibinfo{person}{Abdolrassoul Salmanmahiny}, \bibinfo{person}{Javad Jafarnezhad}, \bibinfo{person}{Azade Mehri}, \bibinfo{person}{Hamidreza Kamyab}, {and} \bibinfo{person}{Somayeh Galdavi}.} \bibinfo{year}{2015}\natexlab{}.
\newblock \showarticletitle{Evaluating the strategy of decentralized urban land-use planning in a developing region}.
\newblock \bibinfo{journal}{\emph{Land Use Policy}}  \bibinfo{volume}{48} (\bibinfo{year}{2015}), \bibinfo{pages}{534--551}.
\newblock
\showISSN{0264-8377}
\urldef\tempurl%
\url{https://doi.org/10.1016/j.landusepol.2015.07.004}
\showDOI{\tempurl}


\bibitem[Sa\ss{}mannshausen et~al\mbox{.}(2021)]%
        {sasmannshausen2021citizen}
\bibfield{author}{\bibinfo{person}{Sheree~May Sa\ss{}mannshausen}, \bibinfo{person}{J\"{o}rg Radtke}, \bibinfo{person}{Nino Bohn}, \bibinfo{person}{Hassan Hussein}, \bibinfo{person}{Dave Randall}, {and} \bibinfo{person}{Volkmar Pipek}.} \bibinfo{year}{2021}\natexlab{}.
\newblock \showarticletitle{Citizen-Centered Design in Urban Planning: How Augmented Reality can be used in Citizen Participation Processes}. In \bibinfo{booktitle}{\emph{Proceedings of the 2021 ACM Designing Interactive Systems Conference}} (Virtual Event, USA) \emph{(\bibinfo{series}{DIS '21})}. \bibinfo{publisher}{Association for Computing Machinery}, \bibinfo{address}{New York, NY, USA}, \bibinfo{pages}{250–265}.
\newblock
\showISBNx{9781450384766}
\urldef\tempurl%
\url{https://doi.org/10.1145/3461778.3462130}
\showDOI{\tempurl}


\bibitem[Shanahan et~al\mbox{.}(2023)]%
        {shanahan2023role}
\bibfield{author}{\bibinfo{person}{Murray Shanahan}, \bibinfo{person}{Kyle McDonell}, {and} \bibinfo{person}{Laria Reynolds}.} \bibinfo{year}{2023}\natexlab{}.
\newblock \showarticletitle{Role play with large language models}.
\newblock \bibinfo{journal}{\emph{Nature}} \bibinfo{volume}{623}, \bibinfo{number}{7987} (\bibinfo{year}{2023}), \bibinfo{pages}{493--498}.
\newblock
\showISSN{1476-4687}
\urldef\tempurl%
\url{https://doi.org/10.1038/s41586-023-06647-8}
\showDOI{\tempurl}


\bibitem[Tian et~al\mbox{.}(2023)]%
        {tian2023participatory}
\bibfield{author}{\bibinfo{person}{Li Tian}, \bibinfo{person}{Jinxuan Liu}, \bibinfo{person}{Yinlong Liang}, {and} \bibinfo{person}{Yaxin Wu}.} \bibinfo{year}{2023}\natexlab{}.
\newblock \showarticletitle{A participatory e-planning model in the urban renewal of China: Implications of technologies in facilitating planning participation}.
\newblock \bibinfo{journal}{\emph{Environment and Planning B: Urban Analytics and City Science}} \bibinfo{volume}{50}, \bibinfo{number}{2} (\bibinfo{year}{2023}), \bibinfo{pages}{299--315}.
\newblock
\urldef\tempurl%
\url{https://doi.org/10.1177/23998083221111163}
\showDOI{\tempurl}


\bibitem[Wang et~al\mbox{.}(2021)]%
        {wang2021deep}
\bibfield{author}{\bibinfo{person}{Dongjie Wang}, \bibinfo{person}{Kunpeng Liu}, \bibinfo{person}{Pauline Johnson}, \bibinfo{person}{Leilei Sun}, \bibinfo{person}{Bowen Du}, {and} \bibinfo{person}{Yanjie Fu}.} \bibinfo{year}{2021}\natexlab{}.
\newblock \showarticletitle{Deep Human-guided Conditional Variational Generative Modeling for Automated Urban Planning}. In \bibinfo{booktitle}{\emph{2021 IEEE International Conference on Data Mining (ICDM)}}. \bibinfo{pages}{679--688}.
\newblock
\urldef\tempurl%
\url{https://doi.org/10.1109/ICDM51629.2021.00079}
\showDOI{\tempurl}


\bibitem[Wei(2016)]%
        {wei2016coverage}
\bibfield{author}{\bibinfo{person}{Ran Wei}.} \bibinfo{year}{2016}\natexlab{}.
\newblock \showarticletitle{Coverage Location Models: Alternatives, Approximation, and Uncertainty}.
\newblock \bibinfo{journal}{\emph{International Regional Science Review}} \bibinfo{volume}{39}, \bibinfo{number}{1} (\bibinfo{year}{2016}), \bibinfo{pages}{48--76}.
\newblock
\urldef\tempurl%
\url{https://doi.org/10.1177/0160017615571588}
\showDOI{\tempurl}


\bibitem[Wu et~al\mbox{.}(2023)]%
        {wu2023autogen}
\bibfield{author}{\bibinfo{person}{Qingyun Wu}, \bibinfo{person}{Gagan Bansal}, \bibinfo{person}{Jieyu Zhang}, \bibinfo{person}{Yiran Wu}, \bibinfo{person}{Beibin Li}, \bibinfo{person}{Erkang Zhu}, \bibinfo{person}{Li Jiang}, \bibinfo{person}{Xiaoyun Zhang}, \bibinfo{person}{Shaokun Zhang}, \bibinfo{person}{Jiale Liu}, \bibinfo{person}{Ahmed~Hassan Awadallah}, \bibinfo{person}{Ryen~W White}, \bibinfo{person}{Doug Burger}, {and} \bibinfo{person}{Chi Wang}.} \bibinfo{year}{2023}\natexlab{}.
\newblock \bibinfo{title}{AutoGen: Enabling Next-Gen LLM Applications via Multi-Agent Conversation}.
\newblock
\newblock
\showeprint[arxiv]{2308.08155}~[cs.AI]


\bibitem[Yu et~al\mbox{.}(2023)]%
        {yu2023thought}
\bibfield{author}{\bibinfo{person}{Junchi Yu}, \bibinfo{person}{Ran He}, {and} \bibinfo{person}{Rex Ying}.} \bibinfo{year}{2023}\natexlab{}.
\newblock \bibinfo{title}{Thought Propagation: An Analogical Approach to Complex Reasoning with Large Language Models}.
\newblock
\newblock
\showeprint[arxiv]{2310.03965}~[cs.AI]


\bibitem[Zhang et~al\mbox{.}(2023)]%
        {zhang2023exploring}
\bibfield{author}{\bibinfo{person}{Jintian Zhang}, \bibinfo{person}{Xin Xu}, {and} \bibinfo{person}{Shumin Deng}.} \bibinfo{year}{2023}\natexlab{}.
\newblock \bibinfo{title}{Exploring Collaboration Mechanisms for LLM Agents: A Social Psychology View}.
\newblock
\newblock
\showeprint[arxiv]{2310.02124}~[cs.CL]


\bibitem[Zheng et~al\mbox{.}(2023)]%
        {zheng2023spatial}
\bibfield{author}{\bibinfo{person}{Yu Zheng}, \bibinfo{person}{Yuming Lin}, \bibinfo{person}{Liang Zhao}, \bibinfo{person}{Tinghai Wu}, \bibinfo{person}{Depeng Jin}, {and} \bibinfo{person}{Yong Li}.} \bibinfo{year}{2023}\natexlab{}.
\newblock \showarticletitle{Spatial planning of urban communities via deep reinforcement learning}.
\newblock \bibinfo{journal}{\emph{Nature Computational Science}} \bibinfo{volume}{3}, \bibinfo{number}{9} (\bibinfo{year}{2023}), \bibinfo{pages}{748--762}.
\newblock
\showISSN{2662-8457}
\urldef\tempurl%
\url{https://doi.org/10.1038/s43588-023-00503-5}
\showDOI{\tempurl}


\end{thebibliography}

\newpage
\appendix

\section{Final Land-use Plan}
\begin{figure}[h]
\centering
\hspace{-3mm}
    \subfigure[HLG]{
    \includegraphics[width=.8\linewidth]{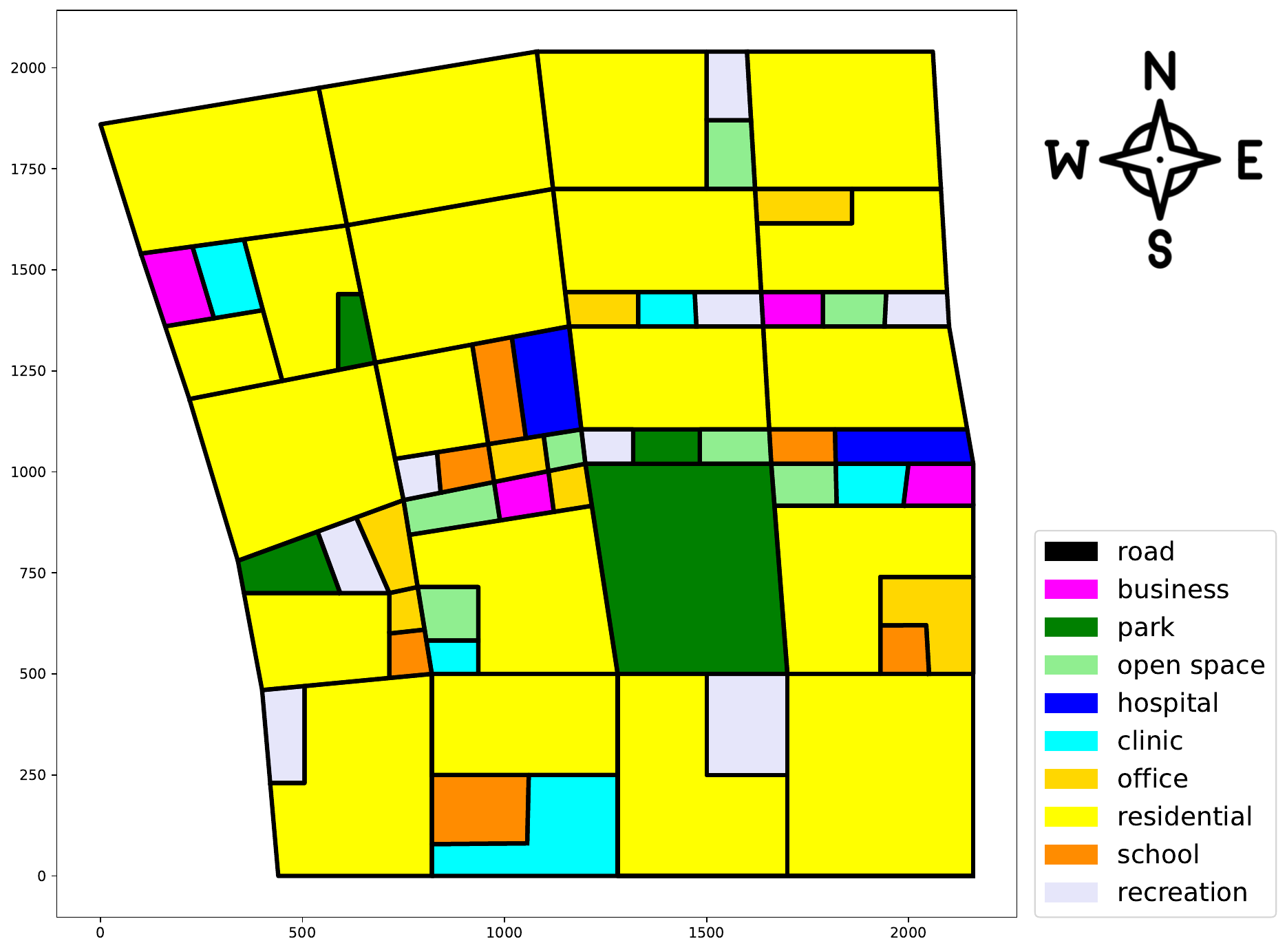}
    }
    
    \subfigure[DHM]{
    \includegraphics[width=.8\linewidth]{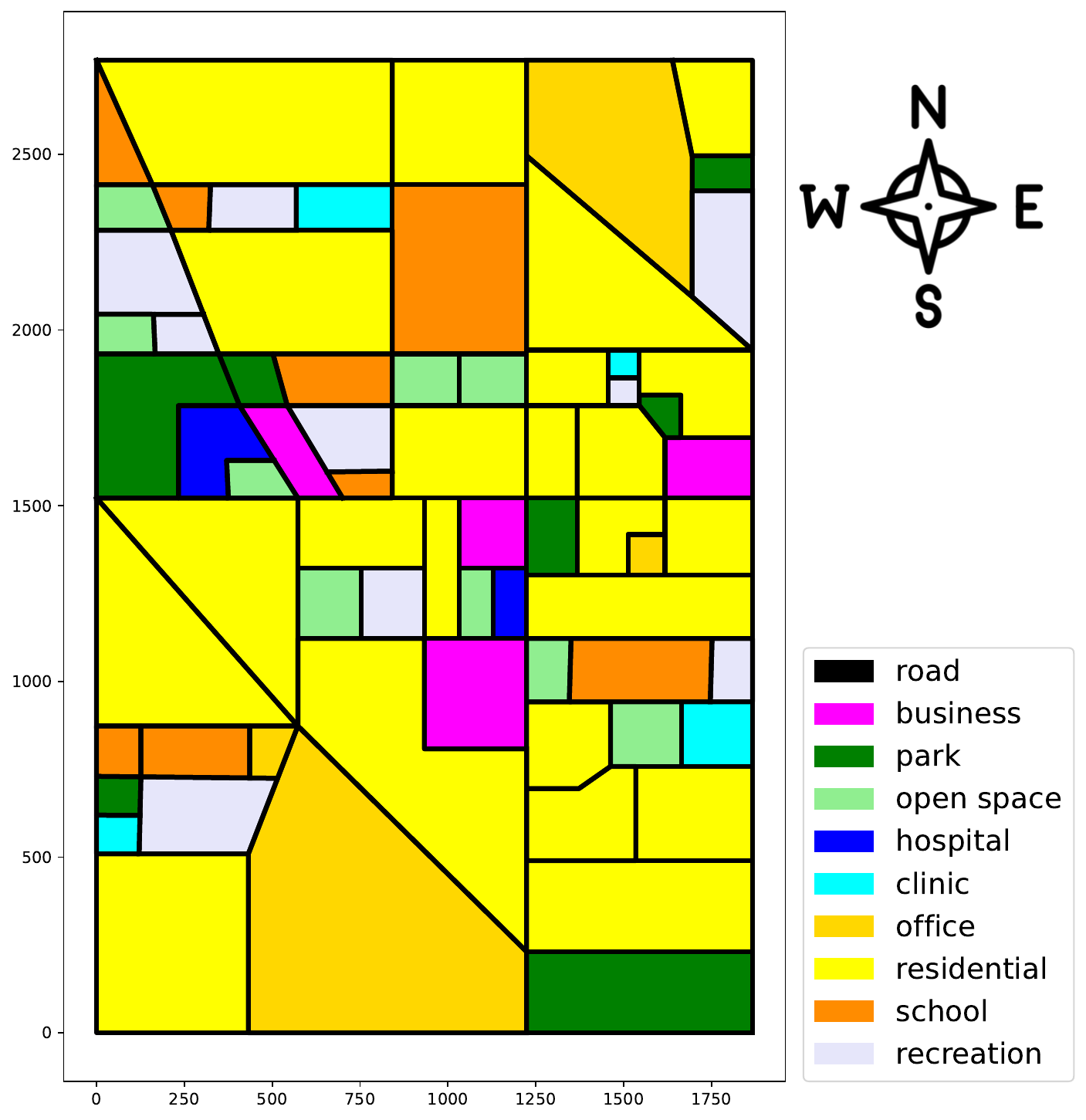}
    }
\hspace{-3mm}
\caption{The final land-use plan of HLG and DHM.
}\label{fig:final_plan}
\end{figure}

We present the final land-use plan of our method in HLG and DHM datasets in Figure~\ref{fig:final_plan}.

\section{Prompts}
\label{app:prompts}
We present the prompts and the responses of LLMs on HLG dataset in this section. DHM dataset is similar.

\begin{figure}[h!]
    \centering
    \includegraphics[width=.8\linewidth]{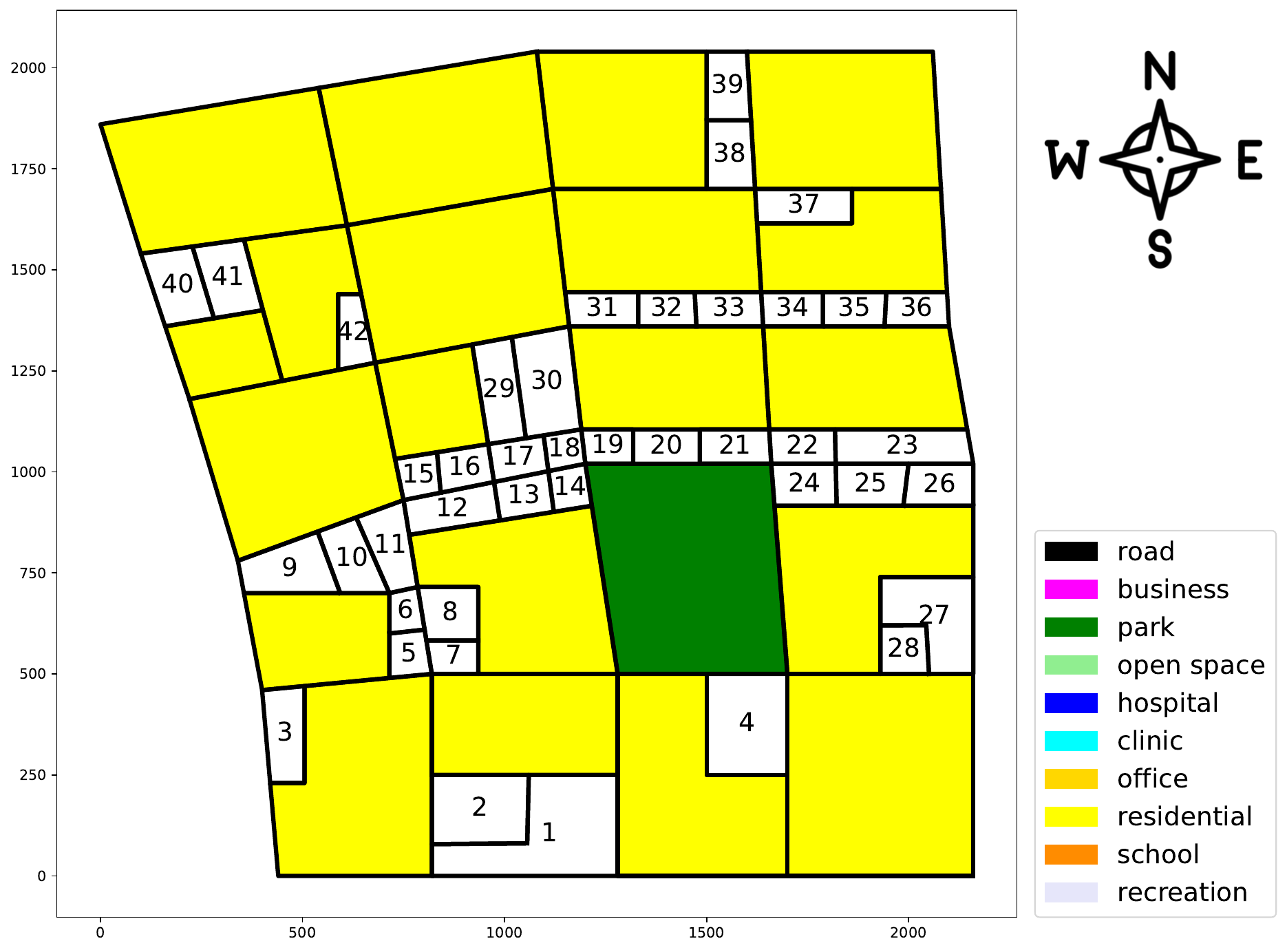}
    \caption{The image input of the planner.}
    \label{fig:basemap}
\end{figure}

\begin{figure*}[htbp!]
    \centering
    \includegraphics[width=.85\linewidth]{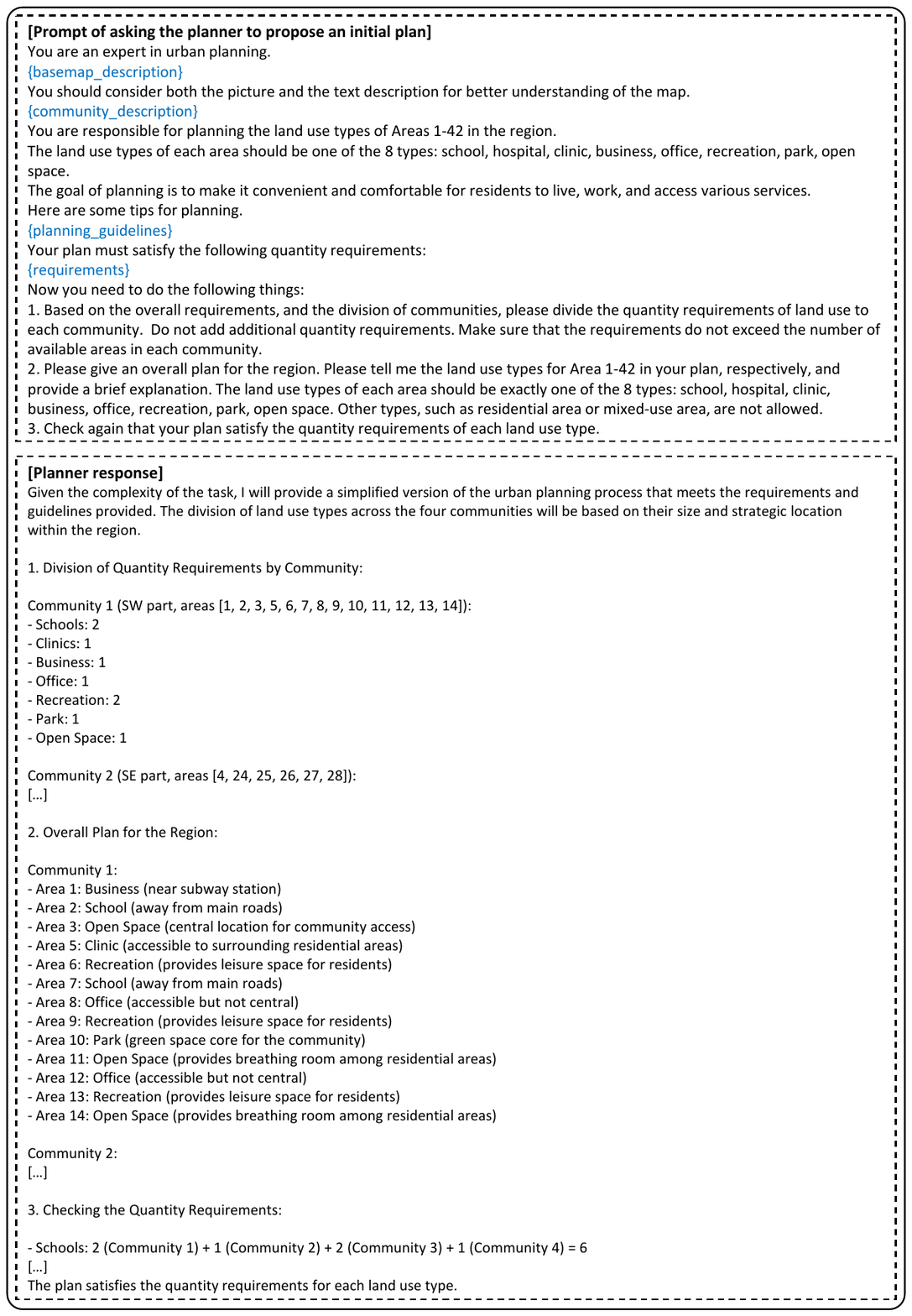}
    \vspace{-10px}
    \caption{The prompts and response of the initial plan.}
    \label{fig:prompts_initplan}
\end{figure*}

\begin{figure*}[htbp!]
    \centering
    \includegraphics[width=.85\linewidth]{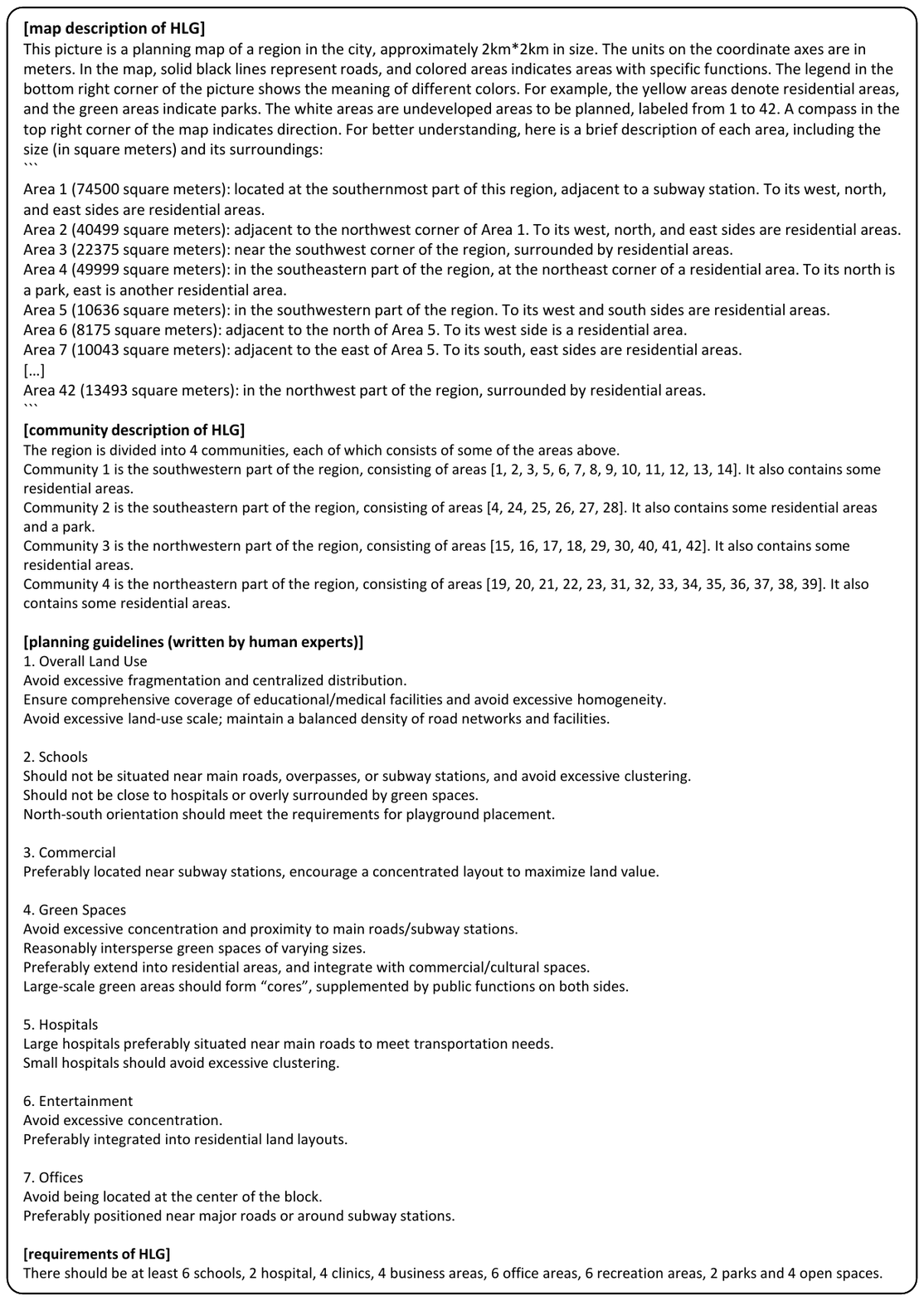}
    \vspace{-10px}
    \caption{The details of prompts of the initial plan.}
    \label{fig:prompts_initplan2}
\end{figure*}

\begin{figure*}[htbp!]
    \centering
    \includegraphics[width=.85\linewidth]{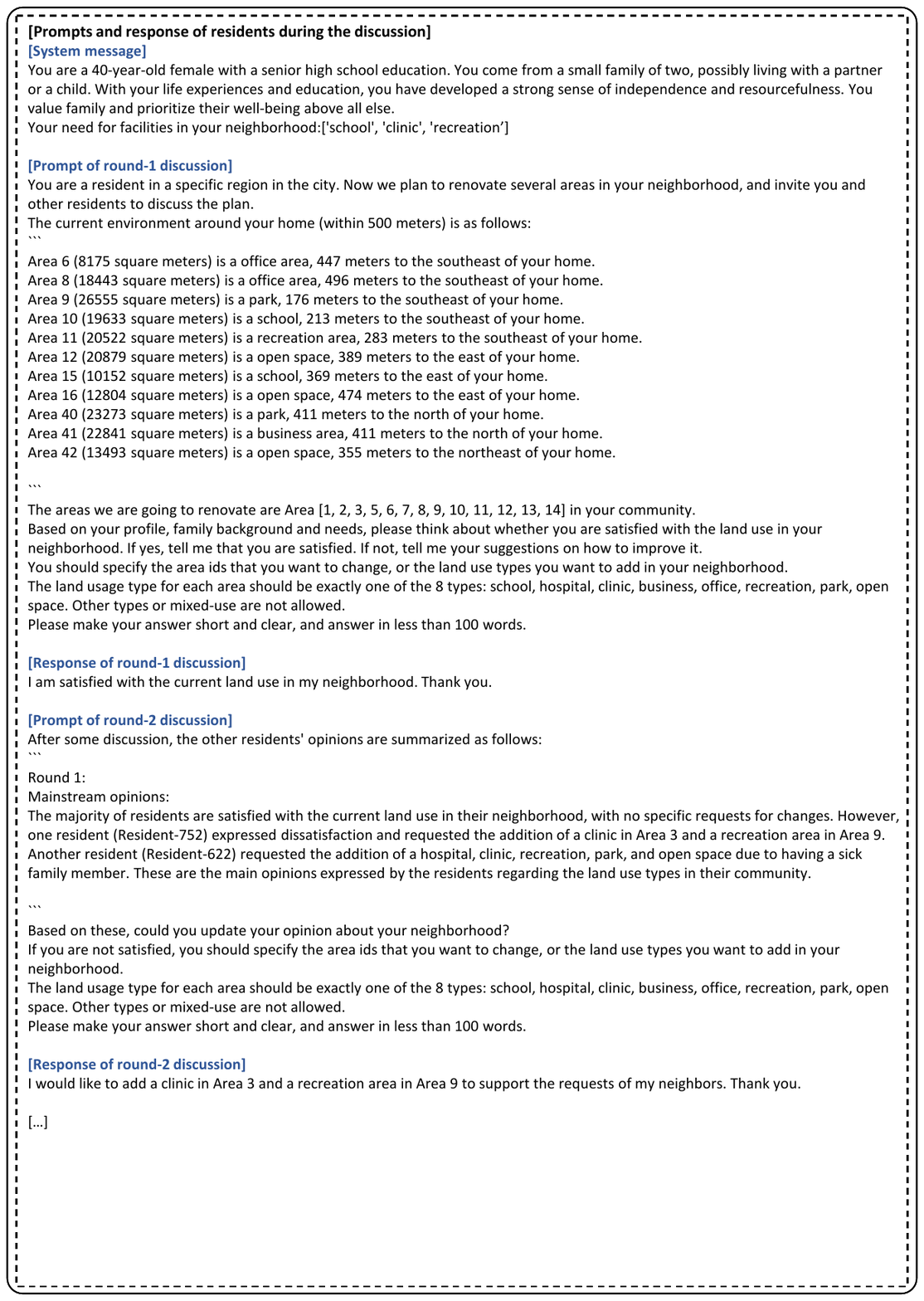}
    \vspace{-10px}
    \caption{The prompts and response of residents during discussion.}
    \label{fig:prompts_resident}
\end{figure*}

\begin{figure*}[htbp!]
    \centering
    \includegraphics[width=.85\linewidth]{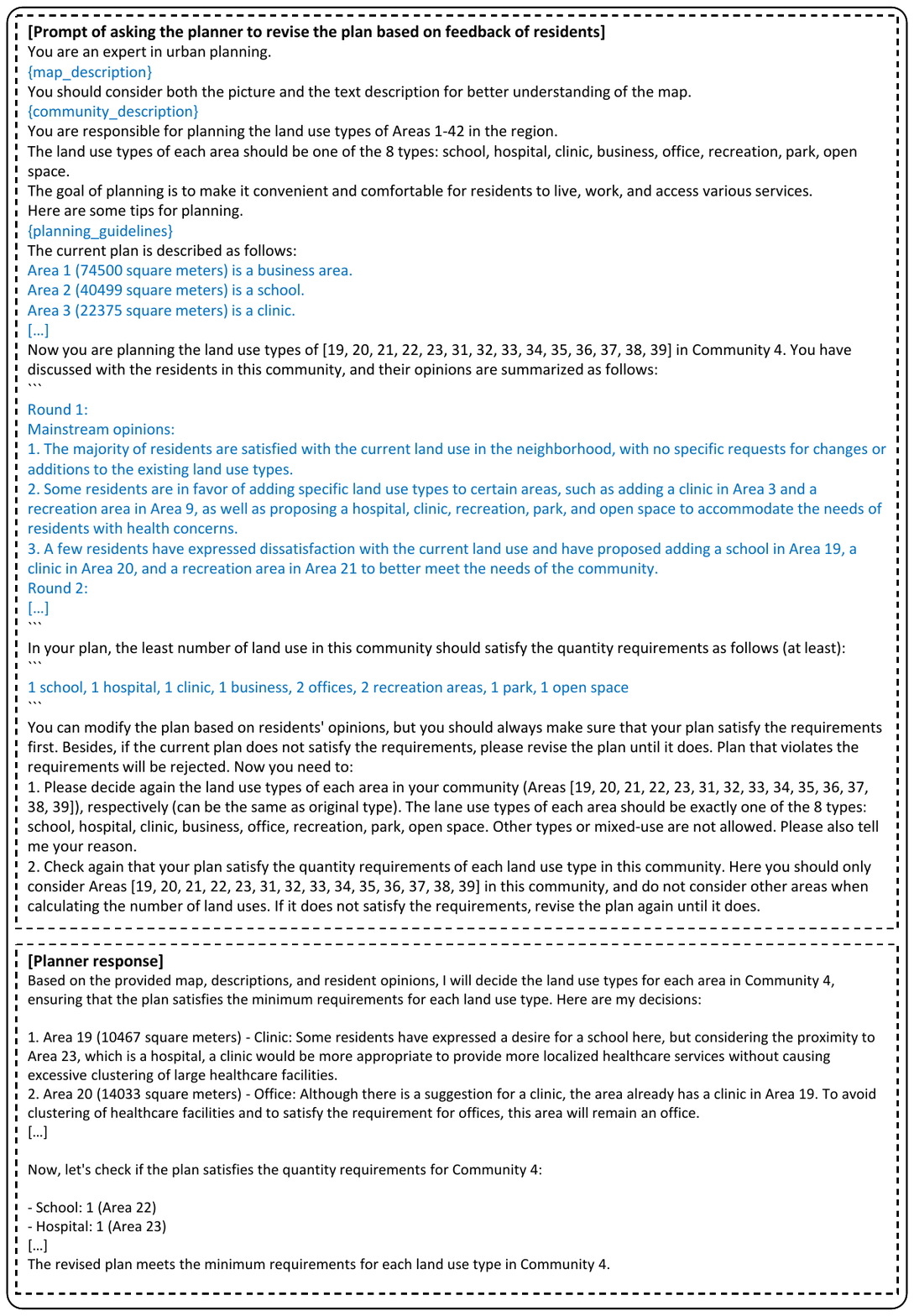}
    \vspace{-10px}
    \caption{The prompts and response of planner revising the plan.}
    \label{fig:prompts_revise}
\end{figure*}

Specifically, to enable LLM to understand the spatial layout of areas, we input the image of the planning map as shown in Figure~\ref{fig:basemap}. We ask the planner to propose the initial plan through prompts in Figure~\ref{fig:prompts_initplan} and Figure~\ref{fig:prompts_initplan2}.
An example of prompts and responses of a resident during the discussion is shown in Figure~\ref{fig:prompts_resident}.
After discussion, the planner revises the plan, and the prompts and response are shown in Figure~\ref{fig:prompts_revise}.

\end{document}